\documentclass{article}
\usepackage{graphicx}
\usepackage{wrapfig}
\usepackage{pgfplots}
\pgfplotsset{compat=1.8}
\usepackage{pgfplotstable}
\usepackage{tikz}
\usetikzlibrary{arrows,shapes,trees,plotmarks}
\usepackage{tabulary}
\usepackage{multirow}
\usepackage[margin=1in]{geometry}
\usepackage{titlesec}
\usepackage[labelfont=bf,textfont=it,font=footnotesize]{caption}
\usepackage{parskip}
\usepackage{array}
\usepackage{enumitem}
\usepackage[super,comma,sort&compress,numbers]{natbib}
\usepackage{booktabs}
\usepackage{listings}
\usepackage{float}
\usepackage{subcaption}
\usepackage{fancyvrb}
\usepackage{verbatim}
\usepackage{todonotes}
\usepackage{url}

\graphicspath{{Figures/}}
\DeclareGraphicsExtensions{.pdf,.png,.jpg}

\titlespacing\section{0pt}{6pt plus 1pt minus 0pt}{3pt plus 1pt minus 0pt}
\titlespacing\subsection{0pt}{4pt plus 1pt minus 0pt}{3pt plus 1pt minus 0pt}
\titlespacing\subsubsection{0pt}{4pt plus 1pt minus 1pt}{3pt plus 1pt minus 1pt}
\titleformat{\section}{\large\bfseries\sffamily}{\thesection}{1em}{}
\titleformat{\subsection}{\normalsize\bfseries\sffamily}{\thesubsection}{1em}{}
\titleformat{\subsubsection}{\normalsize\bfseries\sffamily}{\thesubsubsection}{1em}{}

\usepackage{siunitx}
\usepackage[T1]{fontenc}
\usepackage{mathpazo}

\usepackage{hyperref}
\hypersetup{
    colorlinks,
    linkcolor={magenta},
    citecolor={blue},
    urlcolor={blue!80!black},
    breaklinks=true,
	plainpages=true
}
\newcommand{\cref}[2]{\hyperref[#2]{#1~\ref*{#2}}}

\usepackage{amsmath,amsfonts,bm, amsthm, algorithm, algpseudocode}
\usepackage{nicefrac}        

\newcommand{\colref}[2]{\hyperref[#2]{#1~\ref*{#2}}}
\newcommand{\coloredref}[2]{\hyperref[#2]{#1~\ref*{#2}}}
\newcommand{\coloredsubref}[3]{\hyperref[#2]{#1~\ref*{#2}{#3}}}

\newcommand{\Figref}[1]{\colref{Figure}{#1}}

\newcommand{\Secref}[1]{\colref{Section}{#1}}

\def\eqref#1{eq.~\ref{#1}}

\newcommand{\Eqnref}[1]{\colref{Equation}{#1}}

\newcommand{\Algref}[1]{\colref{Algorithm}{#1}}

\newcommand{\Tabref}[1]{\colref{Table}{#1}}
\def\Tableref#1{Table~\ref{#1}}

\def\ceil#1{\lceil #1 \rceil}

\def\1{\bm{1}}

\DeclareMathAlphabet{\mathsfit}{\encodingdefault}{\sfdefault}{m}{sl}
\SetMathAlphabet{\mathsfit}{bold}{\encodingdefault}{\sfdefault}{bx}{n}

\theoremstyle{plain}

\theoremstyle{remark}

\theoremstyle{definition}

\theoremstyle{plain}

\theoremstyle{plain}

\theoremstyle{definition}

\providecommand{\corollaryname}{Corollary}
\providecommand{\lemmaname}{Lemma}
\providecommand{\problemname}{Problem}
\providecommand{\remarkname}{Remark}
\providecommand{\theoremname}{Theorem}

\newcommand{\mvec}[1]{{\underline{#1}}}

\newcommand{\bunderline}[2][4]{\underline{#2\mkern-#1mu}\mkern#1mu}

\newcommand{\grad}{\bunderline{{\nabla}}}

\newcommand{\partialder}[2]{\frac{\partial #1}{\partial #2}}

\newcommand{\neuralMap}{G_{nn}}

\newcommand{\diffnet}{\textsc{DiffNet}}
\newcommand{\mgdiffnet}{\textsc{MGDiffNet}}
\newcommand{\nel}{n_{el}}

\begin{document}

\begin{center}
{\usefont{OT1}{phv}{b}{sc}\selectfont\Large{Distributed Multigrid Neural Solvers on Megavoxel Domains}}

{\usefont{OT1}{phv}{}{}\selectfont\normalsize
{Aditya Balu$^1$, Sergio Botelho$^2$, Biswajit Khara$^1$, Vinay Rao$^2$, Chinmay Hegde$^3$, 
Soumik Sarkar$^1$\\ Santi Adavani$^2$, Adarsh Krishnamurthy$^1$}, Baskar Ganapathysubramanian$^1$}

{\usefont{OT1}{phv}{}{}\selectfont\normalsize
{$^1$ Iowa State University\\
$^2$ RocketML Inc.\\
$^3$ New York University\\
}}
\end{center}

\begin{figure*}[h!]
    \centering
    \includegraphics[width=0.99\linewidth,trim={0in 0in 0in 0in},clip]{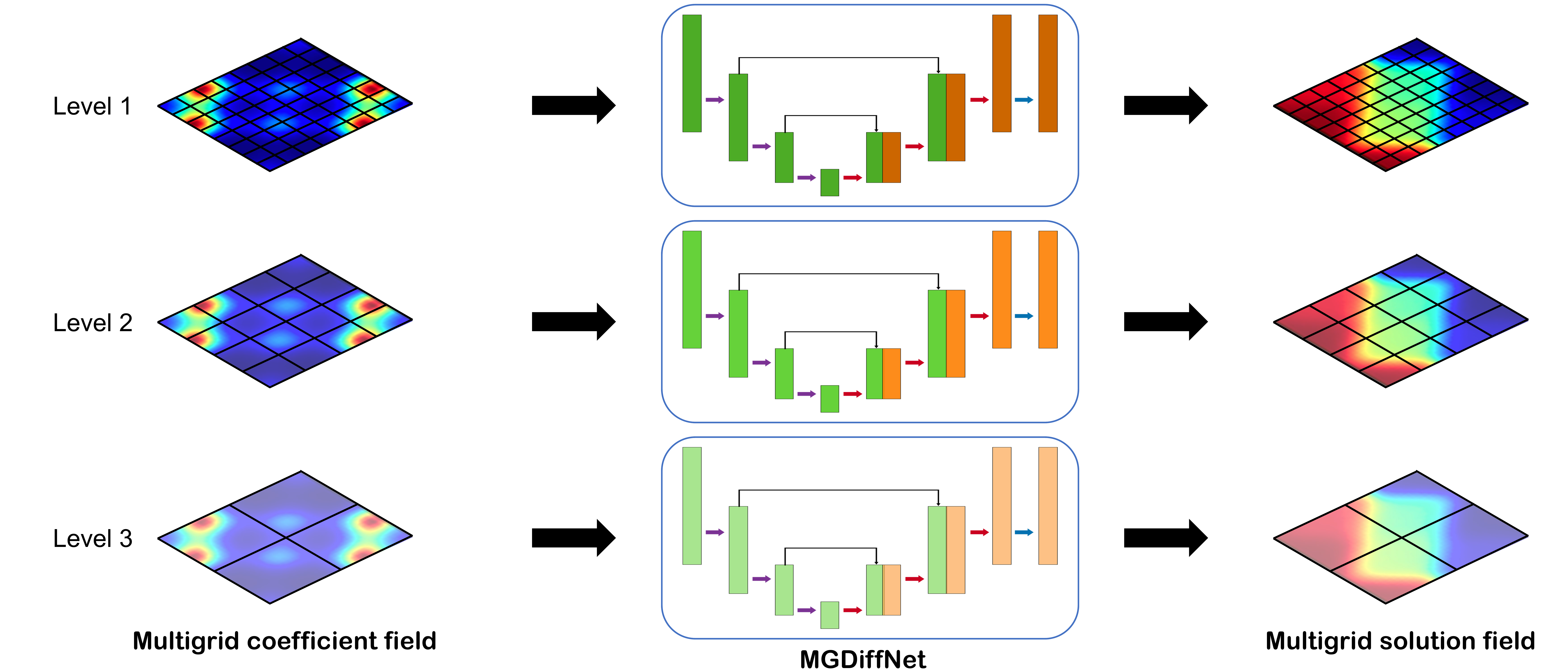}  
    \caption{We demonstrate a distributed multigrid strategy to train a  neural solver that maps a coefficient field with solution field for a given parametric PDE. Coefficient fields at different multigrid resolutions are input to the same underlying network architecture at different stages of training to train the architecture at the highest resolution. }
    \label{fig:overview}
\end{figure*}

\section*{Abstract}

We consider the distributed training of large scale neural networks that serve as PDE solvers producing full field outputs. We specifically consider neural solvers for the generalized 3D Poisson equation over megavoxel domains. A scalable framework is presented that integrates two distinct advances. First, we accelerate training a large model via a method analogous to the multigrid technique used in numerical linear algebra. Here, the network is trained using a hierarchy of increasing resolution inputs in sequence, analogous to the `V’, `W’, `F' and `Half-V' cycles used in multigrid approaches. In conjunction with the multi-grid approach, we implement a distributed deep learning framework which significantly reduces the time to solve. We  show scalability of this approach on both GPU (Azure VMs on Cloud) and CPU clusters (PSC Bridges2). This approach is deployed to train a generalized 3D Poisson solver that scales well to predict output full field solutions up to the resolution of $512\times512\times512$ for a high dimensional family of inputs.

\subsection*{Keywords}

Physics aware neural networks $|$
Distributed training $|$ Multigrid $|$ Neural PDE solvers

\vspace{0.1in}

\section{Introduction}\label{Sec:Introduction}

In recent years, several data-driven~\citep{rudy2019data,tompson2017accelerating} and data-free \citep{raissi2019physics,kharazmi2019variational,sirignano2018dgm,yang2018physics,pang2019fpinns,karumuri2020simulator,han2018solving,michoski2019solving,samaniego2020energy} approaches for solving partial differential equations (PDEs) have been proposed. The backbone of these approaches is the use of (deep) neural networks, which have proven to be capable of learning complex non-linear relationships between the inputs and the outputs. For a subset of these neural PDE solver approaches, the intent is to obtain field predictions, which can then be used to fill in a sparse amount of observable data~\citep{raissi2018hidden,cai2021artificial} or optimize the input parameters for inverse design~\citep{lu2021physics,chen2020physics}. The motivation behind such networks is to have a fast surrogate model that can quickly provide full-field solutions at a much lower cost than traditional numerical simulators. This approach is especially useful in computational design optimization, where hundreds (or thousands) of simulations are necessary to obtain an optimal design, making it computationally expensive or impractical to use traditional scientific simulators. While reduced-order modeling approaches exist for performing such design optimization, they do not necessarily capture the complete complex relationship of the underlying physics. Specifically, for design optimization at very high resolutions, reduced-order modeling may not capture the fine-scale features driving the design figure of merits (for instance, initiation of combustion instabilities). Furthermore, field reconstruction (for instance, for infilling contaminant distributions from sparse measurements) requires fast estimation of the full field. This is the motivation for the current work, where we explore the idea of using neural PDE solvers to obtain the field solutions for parametric PDEs at a very high spatial resolution to enable future computational design at these high resolutions.

A large fraction of neural solvers are designed for pointwise prediction, i.e., the networks in these cases take as input a vector $ \mvec{x}$ of locations in the spatial domain $D$, and produces an output vector $\mvec{u}$, by calculating the value of $u$ at each point. They exploit the ideas of automatic differentiation~\citep{paszke2017automatic} to solve the PDE by minimizing the residual over a set of sampled points $\mvec{x}$. Due to this implicit representation, these methods do not require a mesh and rely on collocating points from the domain randomly. Apart from minimizing the volumetric residual, these approaches also satisfy the prescribed boundary conditions. Some of these methods satisfy/apply the boundary conditions exactly~\citep{lee1990neural,lagaris1998artificial,malek2006numerical}, while others do that in an approximate (weak) sense~\citep{lagaris2000neural,raissi2019physics,sirignano2018dgm}. While the state-of-the-art methods mentioned here show great promise in mapping the complex non-linear relationship between the domain and the field values representing the physics, these methods have the following limitations:
\begin{enumerate}[left=0pt,topsep=0.1in]
    \item \textbf{Need for hyper-parameter tuning:} The methods that approximately satisfy the boundary conditions do so by adding a loss function with respect to the specified boundary conditions. However, the losses have to be carefully weighed, making this a non-trivial exercise in hyper parameter tuning~\citep{van2020optimally}. While recent work like Variational PINN~\citep{kharazmi2021hp}, neurodiffeq~\citep{chen2020neurodiffeq} alleviate this issue (by the exact imposition of boundary conditions, instead of another loss), these are not yet fully developed for arbitrary boundary conditions.
    \item \textbf{Single instance solution:} Most of the approaches above use an implicit representation of the domain where the input are the points $\mvec{x}$ for performing the prediction. Although the implicit representation has several advantages, such as its capability to predict the fields for any arbitrary resolution of points, there are disadvantages, such as the inability to provide topological information about the geometry. Topological information is essential for developing a robust solver that can handle changing the input geometry or the input parameters. Therefore, the above methods suffer from the limitation of their applicability to a single instance of the PDE and do not solve a family of parametric PDE instances. Recent works such as SimNet~\citep{hennigh2020nvidia} attempt to capture a small domain of parametric cases instead of the complete field representation of the parametric PDE.
    \item \textbf{Scalability:} Most of these approaches (although fundamentally scalable) have not been well explored in applications to 3D spatial domains due to computational costs involved in training such deep learning models. With the increase in dimensionality, there is an increase in the number of collocation points sampled (the spatial resolution). Further, enforcing boundary conditions is much more challenging (in weak enforcement of the boundary condition). Apart from these technical issues, computational issues such as the computational cost involved in training these networks are also challenging.
\end{enumerate}

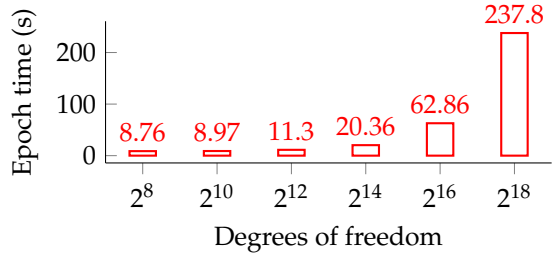
\begin{figure}[!t]
\centering
\begin{minipage}{.7\linewidth}
  \centering
 \begin{tikzpicture}
     \begin{axis}[
         width=0.65\linewidth, 
         height = 0.3\linewidth,
        ybar,
         xmode=log,log basis x={2},
        xtick=data,
        nodes near coords,
        axis y line*=left,
        axis x line*=bottom,
         xlabel= {Degrees of freedom}, 
         ylabel={Epoch time (s)},
         legend style={at={(0.95,0.95)},anchor=north east,legend columns=1}, 
         x tick label style={rotate=0,anchor=north} 
      ]
      \addplot [thick,red]
      table[x expr={\thisrow{resolution}},y expr={\thisrow{time}},col sep=comma]{Data/epochtimings.csv};
     \end{axis}
     \vspace{-20pt}
  \end{tikzpicture}
\end{minipage}
 \caption{Time taken per epoch for performing training at different resolutions of the 2D solution field using same network architecture.}\label{fig:epochtimes}
\end{figure}

A limited number of efforts address these issues. For example, \citet{liao2019deep} resolve application of essential boundary conditions by using Nitsche's variational formulation. \citet{khoo2017solving} extend efforts for solving parametric PDEs. In additiona to these mathematical developments, recent work such as \citet{botelho2020deep}, and \citet{yang2019highly} enable the  scalable training of models used for solving PDEs. Specifically, \citet{yang2019highly} demonstrates the scalability of the framework to 27,500 GPUs. However, the application of these methods in 3-dimensional spatial domains is computationally expensive. As the spatial domain increases, traditional PINN (and its variants) need a vast number of collocation points. Similarly, in the parametric setting, using a convolutional neural network~\citep{botelho2020deep,khoo2017solving}, the voxel resolution creates computational and memory requirement challenges. For example, in \Figref{fig:epochtimes} we see that the computational time per epoch increases quadratically with the increase in the resolution of the spatial domain. These challenges persist, especially for training neural PDE solvers at scale.  

Data-parallel distributed deep-learning strategies are often used to overcome memory limitations, where multiple replicas of a model are simultaneously trained to optimize a single objective function. Typically, universities and government research labs either use on-premise HPC clusters or supercomputers such as the Summit, Bridges2, Frontera, and Stampede2. In this paper, we use a distributed deep learning strategy for performing our training on the Bridges2 cluster running on CPU nodes. However, most of these systems have very few GPU nodes (except for Summit, having 27,360 GPUs). Therefore, we use the Microsoft Azure on-demand HPC virtual machines for performing our distributed experiments on the GPU. This is especially topical, given recent efforts by federal agencies (like the US NSF) for providing cloud access via the CloudBank service.

In addition to using distributed deep learning, we also propose a new training scheme inspired by the multigrid approaches to solving PDEs. The key idea is to use a variational formulation of the loss function to train the neural network at different resolutions or levels (similar to different levels in the multigrid approach). This approach is particularly useful because the training in the lower resolutions is much faster (see \Figref{fig:epochtimes}) than the training time at higher spatial resolutions. We explore strategies for efficient and scalable training of neural PDE solvers based on this approach.

\textbf{Remark}: While our PDE application motivates these developments, the distributed multigrid approach can be used to train any fully convolutional neural network that maps input fields to output fields that are resolution agnostic. This encompasses diverse applications, including semantic segmentation and image-to-image translation prevalent in computer vision. 

The main contributions of this paper are:
\begin{enumerate}[left=0pt,topsep=0.0in]
  \item A variational loss function to solve PDEs (similar to previously proposed ideas~\citep{sirignano2018dgm,liao2019deep,kharazmi2019variational,kharazmi2021hp}) but with the exact application of boundary conditions.
  \item A multigrid-inspired training scheme for training the networks at higher resolutions. We explore several multigrid training schemes and perform a detailed comparison with the direct training of the neural network at high resolutions.
  \item Scaling of the approach to very high resolutions (up to $512\times512\times512$ voxel resolution) using a distributed data-parallel training of large-scale networks in 3D using CPU (on PSC Bridges2) and GPU (on Azure VMs) clusters.
\end{enumerate}

The rest of the paper is arranged as follows: we first explain the mathematical preliminaries in \Secref{sec:math_prelim}; we explain the algorithmic contributions of our work in \Secref{sec:alg_development}; we present the scaling and timing results in \Secref{sec:results}; and finally, we conclude and provide a few remarks on possible future work.

\section{Mathematical preliminaries}
\label{sec:math_prelim}

\subsection{Convolutional Neural Networks (CNNs)}
A deep neural network consists of several layers of connections forming one network, which takes an input $\gamma_{in}$ and produces an output $\gamma_{out}$. Each connecting layer ($l_i$) in the network can be represented as $\gamma_{l_{i+1}} = \sigma(W_{l_i} \cdot \gamma_{l_i} + b_{l_i})$, where $\sigma(...)$ represents a non-linear activation function, $W_{l_i}$ and $b_{l_i}$ are the weights and biases in the connection. The connections could be as simple as a dense connection between every input neuron and output neuron in the layer. However, all connections in a dense connection layer may not be meaningful, and the sample complexity to learn the connections would be high. A convolution connection instead of a dense connection is more efficient for such connections. The convolution operation ($\otimes$) between a $3D$ input representation $\gamma$ and a corresponding $3D$ weight, $W$ is given by
\begin{eqnarray} \label{LAMeq}
W[m,n,p] \otimes \gamma[m,n,p] = \nonumber \,\,\,\,\,\,\,\,\,\,\,\,\,\,\,\,\,\,\,\,\,\,\,\,\,\,\,\,\,\,\,\,\,\,\,\,\,\,\,\,\,\,\,\,\,\,\,\,\,\, \\ 
\sum_{i=-h}^{i=h}\sum_{j=-l}^{j=l}\sum_{k=-q}^{k=q}  W[i,j,k] \cdot \gamma[m-i,n-j,p-k]
\end{eqnarray}

A series of convolutional connections, non-linear activations, and pooling forms a CNN. CNNs are more prevalent in deep learning due to their efficacy in capturing the topological information in datasets such as images, videos, voxels, etc. Several recent papers have utilized such neural networks for producing field predictions~\citep{zhu2018bayesian,zhu2019physics,ranade2021discretizationnet,ozbay2019poisson}. In the next section, we provide details of the network used in this paper. Now, we shall cover some preliminaries for solving PDEs using neural networks.

\subsection{\diffnet{}: Solving PDEs using CNNs}
Consider a bounded open (spatial) domain $D \in \mathbb{R}^n, n\geq 2$ with a Lipschitz continuous boundary $\Gamma = \partial D$. We will denote the domain variable as $\mvec{x}$, where the underbar denotes a vector or tuple of real numbers. In $ \mathbb{R}^n $, we have $ \mvec{x} = (x_1,x_2,\ldots,x_n) $; but for 2D and 3D domains, we will use the more common notation $ \mvec{x} = (x,y) $ and $ \mvec{x} = (x,y,z) $ respectively. On this domain $ D $, we consider an abstract PDE on the function $u: D \rightarrow \mathbb{R}$ as:

\begin{subequations}\label{pde:abstract-representation-continuous}
	\begin{align}
	\mathcal{N}[u; s(\mvec{x},\omega)] &=f(\mvec{x}),\quad \mvec{x}\in D \label{pde:abstract-equation}\\
	\mathcal{B}(u,\mvec{x})&=g(\mvec{x}),\quad \mvec{x}\in\Gamma \label{pde:abstract-bc}
	\end{align}
\end{subequations}
where $\mathcal{N}$ is a differential operator (possibly nonlinear) operating on a function $ u $. The differential equation also depends on the data of the problem $s$ which in turn is a function of the domain variable $ \mvec{x} $ and parameter $ \omega $. Thus $ \mathcal{N} $ is essentially a family of PDE's parameterized by $ \omega $. $ \mathcal{B} $ is a boundary operator acting on $ u $. In general, there can be multiple boundary operators for different parts of the boundary $ \Gamma $.

Given a PDE along with some boundary conditions, such as one presented in \Eqnref{pde:abstract-representation-continuous}, the goal is to find a solution $ u $ that satisfies \Eqnref{pde:abstract-representation-continuous} as accurately as possible. Previous works such as \citep{lagaris1998artificial,raissi2019physics,sirignano2018dgm} seek to find this exact mapping $u: D \rightarrow \mathbb{R}$. But as we present in the next section, we do not have to restrict ourselves to this mapping, and in fact, with the help of deep neural networks coupled with numerical methods, we can find other mappings to retrieve a discrete solution.

In this work, we will focus on the Poisson equation with both Dirichlet and Neumann conditions applied on the boundaries. 

\subsubsection{Poisson Equation: }

Consider the equation:
\begin{align}\label{eq:poisson-pde}
	-\grad \cdot({\nu}(\mvec{x})\grad u) &= f(\mvec{x}) \text{  in  } D
\end{align}
along with the boundary conditions
\begin{align}
	u &= g \text{ on } {\Gamma_D}\label{eq:poisson-bc-dirichlet} \\
	\partialder{u}{n} &= h  \text{ on } {\Gamma_N} \label{eq:poisson-bc-neumann}
\end{align}
where $ \nu $ is the \emph{permeability} (or \emph{diffusivity}), $ f $ is the forcing; $ \Gamma_D $ and $ \Gamma_N $ are the boundaries of the domain $ D $ where Dirichlet and Neumann conditions are specified respectively. We will assume that $ \partial D = \Gamma = \Gamma_D \cup \Gamma_N$. We are mostly interested in a steady-state mass (or heat) transfer through an inhomogeneous medium (material), which means that the material has different properties at different points. The only material property appearing in the Poisson's equation (\Eqnref{eq:poisson-pde}) is $ \nu(\mvec{x}) $, thus the inhomogeneity can be modeled by a spatially varying $ \nu $, i.e., $ \nu = \nu(\mvec{x}) $. Without loss of generality, we consider the following form of the equation:
\begin{align}\label{eq:poisson-kl-intro}
	-\grad\cdot(\tilde{\nu}(\mvec{x})\grad u) &= 0 \text{  in  } D \\
	u(0, y) &= 1 \\
	u(1, y) &= 0 \\
	\partialder{u}{n} &= 0 \text{ on other boundaries} 
\end{align}
where $ D $ is a hypercube domain in $ \mathbb{R}^n $, $ n = 2,3 $.
Here the diffusivity $\tilde{\nu}$ is parametric, and is represented by the following log permeability expression, typically used in geological simulations and in uncertainty quantification:
\begin{align}\label{def:poisson-nu-harmonic-expansion}
	\tilde{\nu}(\mvec{x}; \omega) 
	& = \exp\left(\sum_{i = 1}^{m} \omega_i \lambda_i \xi_i(x)\eta_i(y) \right)
\end{align}
where $\omega_i$ is an $m$-dimensional parameter, $ \lambda $ is a vector of real numbers with monotonically decreasing values in order; and $ \xi $ and $ \eta $ are functions of $ x $ and $ y $ respectively. We take $ m = 4 $, $ \mvec{\omega} = [-3,3]^4 $ and $ {\lambda_i} = \frac{1}{(1+0.25a_i^2)}$, where $ \mvec{a} = (1.72, 4.05, 6.85, 9.82)$. Also $ \xi_i(x) = \frac{a_i}{2}\cos(a_i x) + \sin(a_i x) $ and $ \eta(y) = \frac{a_i}{2}\cos(a_i y) + \sin(a_i y) $.

\subsection{Geometric Multigrid approach}
The geometric multigrid (GMG) is a powerful tool used for scalable numerical linear algebra. The GMG approach works by defining a hierarchy of meshes and sequentially projecting and solving the PDE on these meshes. The advantage of GMG lies in accessing the different regions of the error spectrum of a numerical operator by projecting the error on meshes of varying refinement. This is a powerful concept that can be naturally extended to training CNNs. 

\begin{figure}[h!]
    \centering
    \includegraphics[width=0.99\linewidth,trim={0.2in 0.4in 0.2in 0.4in},clip]{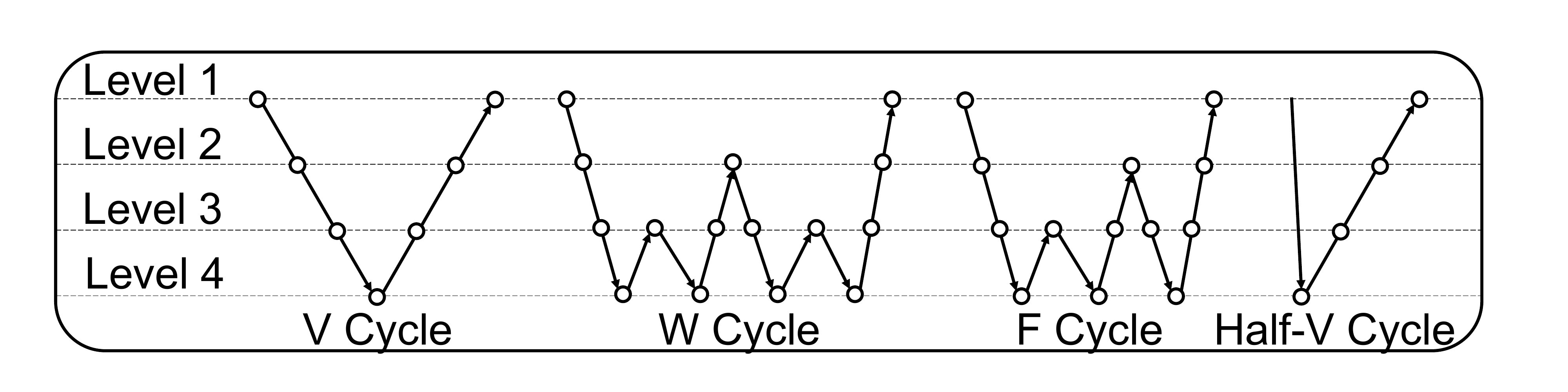}
    \caption{Different multigrid strategies.}
    \label{fig:cycles_desc}
\end{figure}

In GMG, every time the grid is coarsened (as seen in \Figref{fig:cycles_desc}, where the levels indicate increasingly coarser meshes, with Level 1 being the most refined mesh), a range of low frequency of errors from the previous refined mesh are converted to high-frequency errors. On this iteration of coarsening, the fresh set of high-frequency errors are obtained and smoothed. The coarsening of the grid and interpolation onto the coarse grid is called restriction. This idea of segmenting and smoothing the error spectrum of a PDE operator allows for an efficient reduction in errors and parallel scalability. After smoothing the errors at the coarsest level, one can reconstruct the solution by progressively interpolating the solution to finer and finer meshes until one reaches the original resolution. The correction and interpolation from a coarse grid to the finer grid is called prolongation.

Multiple grid hierarchies (or GMG cycles) are used. \Figref{fig:cycles_desc} illustrates some common grid hierarchies in the multigrid approach. It is important to note that solving the system on progressively coarser grids becomes progressively cheaper. In a V-cycle hierarchy, restriction and smoothing are performed until the coarsest grid, and then the prolongation and correction are performed until one reaches the starting mesh resolution. In a W-cycle (second from left in \Figref{fig:cycles_desc}) after restriction and smoothing to the coarsest cycle. However, instead of performing prolongation and correction to the initial mesh resolution, prolongation and correction are used alternatively to minimize the low-frequency errors and improve stability. It is important to note that this does not compromise efficiency as these alternate operations are done on really cheap coarse meshes. Subsequently, correction and prolongation are performed fully to the initial mesh resolution, just like in the V-cycle. The extra expense of the W-cycle compared to the V-cycle is progressively lower for increasing spatial dimensions~\citep{Hackbusch2013}. The F-cycle falls somewhere between V-cycle and W-cycle in terms of expense. It starts with the restriction to the coarsest grid like the V-cycle. After having reached each level the first time, a restriction to the coarsest grid is performed in the prolongation process. The half-V-cycle is a special case of the V-cycle, in which no smoothing is done before the coarsest grid layer. 

In the context of~\mgdiffnet{}, several works have been performed in the context of relating multigrid approaches to deep learning~\citep{wu2020multigrid,chen2020meta,ke2017multigrid} and deep learning approaches to improve multigrid operations~\citep{katrutsa2017deep,luz2020learning,margenberg2020structure,huang2021learning}. Here, we leverage the multigrid hierarchy and try to establish a mapping between the domain and the solution using CNN on every grid layer. Further, a solution from mapping on each grid layer can be used to correct and prolongate to progressively finer mesh resolutions. However, careful scaling and timing analysis is required to determine which of these grid hierarchies provides good scalability while not compromising accuracy. We report this analysis and results in \Secref{sec:results}.

\section{Algorithmic Developments}\label{sec:alg_development}
\subsection{Multigrid Approaches}
We seek a mapping between the input $s$ and the full field solution $u$ in the discrete spaces. $S^d$ denotes the discrete representation of the known quantity $s$. $S^d$ could be either available only at discrete points (perhaps from some experimental data). In many cases, $s$ might be known in a functional form, and thus $S^d$ will be simply the values of $s$  evaluated on the discrete points. Therefore, if we denote a \mgdiffnet{} network by $ \neuralMap$, then $\neuralMap$ takes as input a discrete or functional representation of $s$ and predicts a discrete solution field $U^d_\theta$, where $\theta$ denotes the network parameters. For example, if we consider a PDE defined on a 2D bounded domain, $ \neuralMap $ takes a 2D matrix containing the values of $ s $ and predicts the solution field $ U^d_\theta $ which is also a 2D matrix (as illustrated in \Figref{fig:overview}).

The weights of the network $\neuralMap$ are initialized randomly in the beginning and using optimization schemes, we obtain the network parameters $\theta$, which maps the input coefficients field $s$ to solution field $u$. A first step is designing the loss function based on the finite element method (FEM). 

\subsubsection{FEM Loss:}

The FEM loss involves the weakening of the PDE using an appropriate weighting functions. Let the set $ \mvec{X} = (\mvec{x}_1,\mvec{x}_2,.., \mvec{x}_N) \in \mathbb{R}^{n\times N} $ denote a collection of points in $ \mathbb{R}^n $ that produces a (uniform) discretization of D with a set of non-overlapping elements denoted by $ Q_i $, $ i = 1,2,\ldots,\nel $ such that $ \cup_{i}^{nel} Q_i = D$. we define $ S_i = s(\mvec{x}_i) $ and $ U_i $ an approximation of the unknown $ u(\mvec{x}_i) $.  The unknown solution can be approximated as:
\begin{align}\label{def:fem-function-approximation}
	u^h_\theta = \sum_{i = 1}^{N} \phi_i(\mvec{x})(U_i)_\theta
\end{align}
where $\phi_i$ are the finite element basis functions.

This approximation is plugged into the PDE, after which we invoke Galerkin's method. We multiply the PDE with a test function and reduce the differentiability requirement on $ u^h $ using integration by parts:\footnote{For completeness, we assume $u^h_{\theta} \in V \subset H^1(D)$ where $ H^1(D) $ denotes the Hilbert space of functions on $ D $ that have square integrable first derivatives.}
\begin{align}
	\int_{\Omega} v\left[ \mathcal{N}(u^h_{\theta};s) -f \right] d\mvec{x} &= 0 \text{ }\forall v\in V,
\end{align}
which results in this following (standard FEM) form
\begin{align}
	B(v,u^h_{\theta}) - L(v) &= 0\text{ }\forall v\in V,
\end{align}
where $B(v,u^h_{\theta})$ is the bilinear form that encodes the PDE, while $L(v)$ is the linear form that encodes the load and the boundary conditions. By choosing the test function to be the (unknown) solution, $u^h_{\theta}$, we get an energy functional whose minima is the solution:
\begin{align}\label{eq:loss}
	J(u^h_{\theta}) = \frac{1}{2} B(u^h_{\theta},u^h_{\theta}) - L(u^h_{\theta}).
\end{align}
This energy functional accounts for the PDE as well as all Neumann (and Robin) boundary conditions. This energy functional also serves as our loss function.

\subsubsection{Multigrid Training of \mgdiffnet{}:}

We first define the neural network, $\neuralMap$, to be a fully convolutional neural network with the following properties: 
\begin{enumerate}
    \item the connections between each layer only use convolution (and/or transpose convolution) operations; 
    \item  the downsampling (performed using max-pooling or convolution with stride $>1$) is always a factor of two;
    \item appropriate padding is performed to ward off fence effects.
\end{enumerate}
We also assume that the network architecture has multiple filters in each layer to sufficiently learn features at all levels. Such a neural network, $\neuralMap$, has a significant advantage in performing multigrid training in \mgdiffnet{}. Recall that the filter weights $W$ for a convolution operation are not dependent on the input resolution $d$ and can be used to extract local information from any resolution. Stacking several such convolutional layers (with non-linear pointwise activation functions) provides a fully convolutional network $\neuralMap$.

\begin{algorithm}[!h]
	\caption{Algorithm for \mgdiffnet{}}\label{alg:pde-parametric}
	\begin{algorithmic}[1]
		\Require $ S^d $, $ (U^d_\theta)_{bc} $, $ \alpha $ and \textsc{tol}
		\State Initialize $ \neuralMap $		
		\For{\texttt{epoch} $ \leftarrow $ 1 to \texttt{max\_epoch}}
		\For{\texttt{mb} $ \leftarrow $ 1 to \texttt{max\_mini\_batches}}
		\State Sample $S^d_{mb}$ from the set\\
		\State $(U^d_\theta)_{int,mb} \gets G_{nn}(S^d_{mb})$\\
		\Comment{``int" stands for interior nodes}
		\State $(U^d_\theta)_{mb} \gets (U^d_\theta)_{int,mb}\chi_{int} + (U^d_\theta)_{bc}\chi_{b}$
		\State  $loss_{mb} = L(U^d_\theta)  $
		\State $ \theta \gets optimizer(\theta,\alpha,\grad_\theta(loss_{mb})) $
		\EndFor
		\EndFor
	\end{algorithmic}
\end{algorithm}

Constructing such a network is not difficult. A standard fully convolutional neural network, called U-Net~\citep{ronneberger2015u,cciccek20163d}, satisfies all the requirements mentioned above\footnote{Interestingly, while writing this paper, we came across work that hypothesized deep mathematical connections between numerical methods and neural nets~\citep{alt2021translating}, with a specific call out to a link between multigrid approaches with U-Net architectures. Our work anecdotally validates these assertions.}. The primary use of such as fully convolutional neural network is that the network architecture remains the same for different input resolutions. This means that for learning a smooth solution field, we can perform training of $\neuralMap$ at different resolutions where the network's parameters learn the mapping between the solution field $u$ and the coefficients field $s$. Training will follow the standard use of stochastic gradient-based optimizers (SGD and its variants) as explained in \Algref{alg:pde-parametric}.

\pagebreak 

Here, we consider different multigrid strategies for performing the training (adapted from the traditional multigrid approaches). First, we define different levels of discretization of the domain $D$ where the number of elements in each level are $\nel$, $\nicefrac{\nel}{2}$, $\nicefrac{\nel}{4}$, $\cdots$ in each dimension (i.e. total number of elements is $\nel^2$ in 2D and $\nel^3$ in 3D spatial domains). In \Figref{fig:cycles_desc}, we show different strategies we used to perform multigrid training. The first strategy is to train the network for a few epochs on the finest resolution (level 1) and use the same weights for training on a coarser grid resolution. At this point, the network weights have learned the information at a higher resolution, and now moving to the lower resolution will help learn the neighborhood information of the network at this resolution. This accelerates the training by a large factor at the lower resolutions. The same process continues until reaching the lowest resolution (level 4), where the network is trained until the loss plateaus, or the solution converges. Practically, we use an early-stopping criterion to keep track of when to stop the network training. 

Once the network is trained, due to the fully convolutional neural network, the forward pass of the coefficients through the network itself becomes an excellent starting point for performing interpolation and solving the PDE at a higher resolution (i.e., for a network trained on lower resolution can naturally be used for interpolating on a higher resolution grid).  We now train the network until convergence (defined by the early stopping criteria) to proceed to higher resolutions. The most straightforward strategies are the V cycle and Half-V cycle, where one proceeds from higher to lower and then lower to higher resolutions. However, in the W cycle and F cycle, we perform additional training on other intermediate resolutions to make the network more robust to different scales of the multigrid. In the context of deep learning, these cycles help the network become robust to different resolutions and can learn the unique mapping at all the resolutions. Here, we note that this is only true when the network learning capacity is infinite. Different filters of the convolution operation learn neighborhood information at different scales of the multigrid, thus solving the PDE faster.

\begin{figure*}[!b]
  \centering
  \includegraphics[width=0.99\linewidth,trim={0in 1.5in 0in 1.5in},clip]{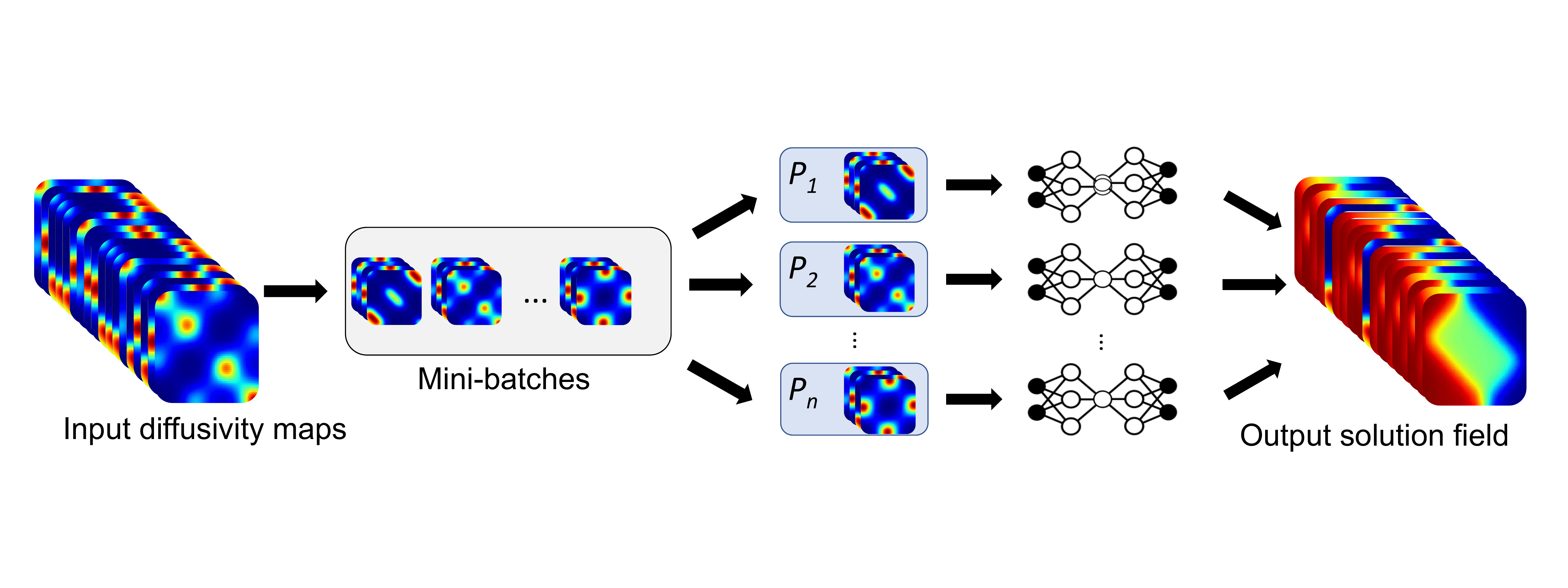}
  \caption{Data-parallel distributed deep-learning: multiple replicas of the model are asynchronously trained by workers, each processing a local subset of the global mini-batch.}
  \label{fig:data_parallel_batch_split}
\end{figure*}

In this study, we only consider one `cycle' of multigrid. While it is certainly possible to extend this for several `cycles' of multigrid and with more variations on which cycle to apply at which stage of the training, we restrict ourselves to just one cycle where each step of the cycle involves longer training time for several epochs. This avoids the problem of moving target (often quoted in relationship with reinforcement learning) where the distribution (or the frequencies of information) of data learned keeps changing, not allowing the network to be properly trained. Further, while the study can be performed at any arbitrary number of multigrid levels, we restrict ourselves to a maximum of 4 levels. Further, all the multigrid prolongation steps are until we reach convergence (defined using an early-stopping criterion). At the same time, all the restriction steps are trained for a fixed number of epochs (because the convergence is not necessary at the higher resolutions in the beginning). Now, we will discuss our distributed data-parallel deep learning implementation.

\subsection{Distributed Deep Learning}

One of the most widely used techniques for performing distributed deep-learning training is the {\it data parallel} strategy, in which identical copies of the model are simultaneously trained by independent processes that work together to minimize a common objective function \citep{bennun_parallel_dnn}. For this to be possible, the training data samples (and their corresponding labels in supervised learning) must be equally split among the workers. Since stochastic optimization-based training already entails splitting the data into mini-batches, this means one has to further split the mini-batches into {\it local} mini-batches, which are then asynchronously processed via forward and back-propagation steps. Local gradients are computed by each worker and collectively averaged using an {\it all-reduce} operation. Once each worker possesses the global gradient vector, they invoke the optimizer to update their local network parameters, which are now in sync with every other worker (see \Figref{fig:data_parallel_batch_split}).

\begin{figure}[!t]
  \centering
  \includegraphics[width=0.6\linewidth,trim={0in 0in 0in 0in},clip]{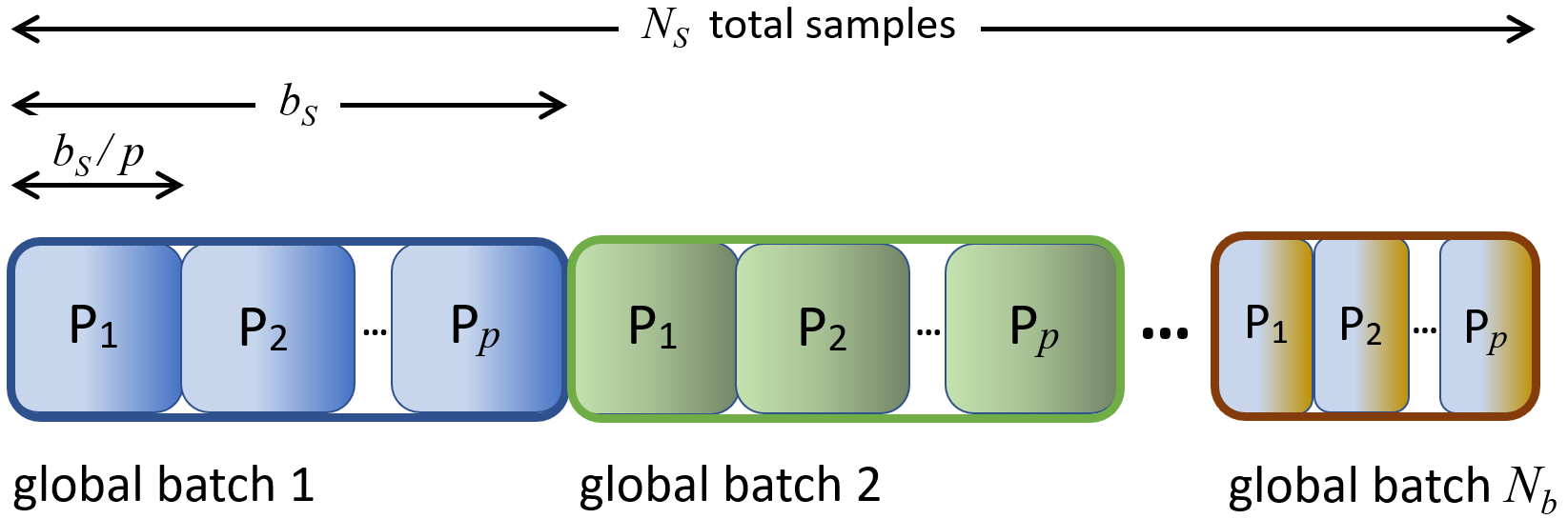}
  \caption{Data splitting across workers in a parallel run: local mini-batches are guaranteed to always have identical sizes at any given time, promoting optimal load balance.}
  \label{fig:mini_batch_split}
\end{figure}

\begin{figure}[t!]
  \centering
  \includegraphics[width=0.55\linewidth]{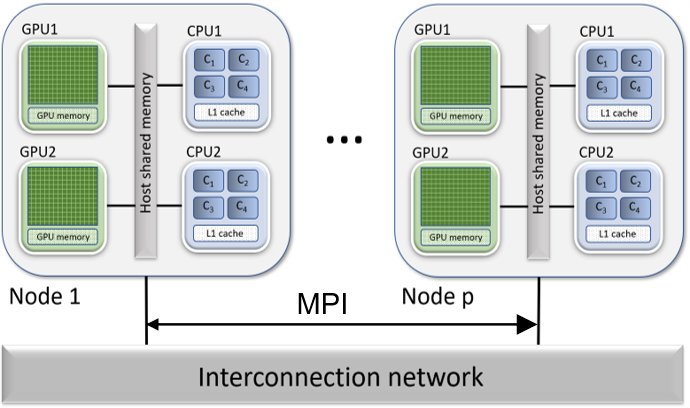}
  \caption{Process-to-process hybrid distribution paradigm: processes communicate via MPI and spawn local threads that exploit intra-node parallelism. \label{fig:deepfusion_hybrid_arch}}
\end{figure}

However, we must ensure that results are independent of the number of workers utilized, an essential tenet of high-performance computing. To accomplish that, we start by augmenting the dataset to make the total number of training samples $N_s$ divisible by the number of workers $p$. Then, each global mini-batch of size $b_s$ is divided into $p$ equal parts, which become the local mini-batches to be dispatched to the $p$ workers, as shown in \Figref{fig:mini_batch_split}. This ensures that the union of the $n^\textrm{th}$ local mini-batches across all workers will be identical to the $n^\textrm{th}$ (global) mini-batch of the corresponding single-processor run,
\begin{align}
\bigcup_{i=0}^{p} (\textrm{LMB})_{n}^{i} = (\textrm{GMB})_{n}
\end{align}
for all $n \in [0, N_b]$, where $N_b = \ceil{N_s / b_s}$ is the number of mini-batches in each training epoch. Module rounding errors during gradient communication, the above scheme thus guarantees that the solution will be independent of the number of workers. It also follows from the arithmetic that, for any global mini-batch size $b_s$ chosen, the local mini-batches processed by workers at any given time will have the same size, thus optimizing load balance.

\begin{table*}[b!]
    \setlength\extrarowheight{2.5pt}
    \caption{Comparison between different multigrid strategies for different resolutions in 2D and 3D.}
    \begin{center}
    \label{Table:strategies}
    \resizebox{0.98\linewidth}{!}{
      \begin{tabular}{|l | l | l | c | r | r | r | r | r |}
        \hline 
        \textbf{Dimension}     &   \textbf{Resolution}     &    \textbf{Strategy}  &   \textbf{Levels}  &   \textbf{Base Time (s)} &   \textbf{MG Time (s)} & \textbf{Base Loss} & \textbf{MG Loss}    &    \textbf{Speedup}\\
        \hline
          \multirow{18}{*}{2D}    &   \multirow{7}{*}{$128\times128$}   &     \multirow{2}{*}{V Cycle} & 3 & \multirow{7}{*}{3021.05}    &  1934.305 & \multirow{7}{*}{0.0510} & 0.0571  & 1.56$\times$ \\
        \cline{4-4}\cline{6-6}\cline{8-9}
          &   &  & 4 &  &  2401.070 &   & 0.0570  & 1.26$\times$ \\
        \cline{3-4}\cline{6-6}\cline{8-9}
          &   & \multirow{2}{*}{Half-V Cycle}  &  3  &  &  3133.861 &  & 0.0568 &  0.96$\times$\\
        \cline{4-4}\cline{6-6}\cline{8-9}
          &   &  &  4  &  & 3275.405  &  & 0.0588 &  0.92$\times$\\
        \cline{3-4}\cline{6-6}\cline{8-9}
          &   & \multirow{2}{*}{W Cycle}  &  3  &   &  2023.778 &  & 0.0569 &  1.49$\times$\\
        \cline{4-4}\cline{6-6}\cline{8-9}
          &   &   &  4  &   & 2512.113 &  & 0.0597 &  1.20$\times$\\
        \cline{3-4}\cline{6-6}\cline{8-9}
          &   & F Cycle  &  4  &  & 2578.451 &  & 0.0584 &  1.17$\times$\\
        \cline{2-9}
          & \multirow{7}{*}{$256\times256$}  &  \multirow{2}{*}{V Cycle} & 3 & \multirow{7}{*}{9248.44}  & 3297.706  & \multirow{7}{*}{0.0165}  & 0.0210  & 2.80$\times$\\
        \cline{4-4}\cline{6-6}\cline{8-9}
          &   &  &  4  &  & 3639.291 &  & 0.0209 &   2.54$\times$\\
        \cline{3-4}\cline{6-6}\cline{8-9}
          &   & \multirow{2}{*}{Half-V Cycle}  &  3  &  & 4585.830 &  & 0.0181 &   2.02$\times$\\
        \cline{4-4}\cline{6-6}\cline{8-9}
          &   &   &  4  &  & 4722.950 &  & 0.0174 &   1.96$\times$\\
        \cline{3-4}\cline{6-6}\cline{8-9}
          &   & \multirow{2}{*}{W Cycle} &  3  &  & 5791.277 &  & 0.0174& 1.60$\times$\\
        \cline{4-4}\cline{6-6}\cline{8-9}
          &   &  &  4  &  & 5597.503 &  & 0.0188 &   1.65$\times$\\
        \cline{3-4}\cline{6-6}\cline{8-9}
          &   & F Cycle &  4  &   & 7401.254 &   & 0.0164&   1.25$\times$\\
        \cline{2-9}
          & \multirow{4}{*}{$512\times512$}  &  V Cycle& 4 &  \multirow{4}{*}{21860.50} &  10352.543 &  \multirow{4}{*}{0.0050} &  0.0058 & 2.11$\times$\\
        \cline{3-4}\cline{6-6}\cline{8-9}
          &   & Half-V Cycle  &  4  &   & 11282.420 &  & 0.0053&   1.94$\times$\\
        \cline{3-4}\cline{6-6}\cline{8-9}
          &   & W Cycle  &  4  &  & 10996.353 &  & 0.0062&  1.99$\times$\\
        \cline{3-4}\cline{6-6}\cline{8-9}
          &   & F Cycle &  4  &  & 17409.934 &  & 0.0053 &  1.26$\times$\\
        \hline
        3D    &   $128\times128\times128$   & Half-V Cycle & 3 & 42422.50 & 7025.314 & 0.0400 & 0.0400 & 6.04$\times$ \\
        \hline 
      \end{tabular}
    }
    \end{center}
\end{table*}


    
Our parallelization strategy leverages both distributed-memory MPI-based communication primitives that handle data transfer across processes, and shared-memory OpenMP or CUDA-based multi-threading that exploits parallelism within a node. This combination of shared memory and message-passing paradigms within the same application is known as {\it hybrid programming} \cite{hybrid_paradigm}, and is illustrated in \Figref{fig:deepfusion_hybrid_arch}. In the specific case of our deep-learning software, MPI collective {\it all-reduce} calls are invoked to handle gradient communication and averaging across workers. They make use of the {\it ring-allReduce} algorithm \cite{sergeev_horovod}, which has a complexity of $O(N_w + log(p))$, where $N_w$ is the number of model parameters. Since $N_w \gg p$, we expect the communication complexity to be almost independent of the cluster size. On the other hand, the engines we use internally to execute forward and back-propagation can spawn their own Open-MP or CUDA threads, which communicate only with other threads within the same MPI process. Since MPI communication only happens outside critical multi-threaded regions, our parallelization strategy can be said to model the {\it process-to-process} hybrid paradigm. The number of processes launched per node and the maximum number of threads spawned by each process will depend on the specs of the cluster and details of the experiment and are chosen in such a way as to maximize resource utilization, minimize communication overhead and fulfill memory requirements.

\section{Results and Discussion}\label{sec:results}
One of the key outcomes of our experiments was to demonstrate a practical approach to train \mgdiffnet{} on domain sizes up to $512^3$. We applied our framework to train \mgdiffnet{} for resolutions up to $256^3$ on GPU-based HPC clusters using on-demand multi-GPU virtual machines on \textbf{Microsoft Azure}. To train DiffNet for resolutions $> 256^3$ we used PSC \textbf{Bridges2} HPC cluster with bare-metal access to CPU nodes. In \Tabref{tab:azure_vs_bridges_specs}, we provide all relevant specifications for Azure and Bridges2 used in our experiments. We first talk about our experiments to study the multigrid approach and then the scaling studies using distributed deep learning.

\begin{table*}[b]
    \setlength\extrarowheight{5pt}
    \small
    \caption{Network Adaptation Studies}
    \begin{center}
    \label{Table:adaptation}
    \begin{tabular}{|l | r | r | r | r | r |}
        \hline 
        \textbf{Strategy}  &    \textbf{Base Time (s)} &   \textbf{MG Time (s)} & \textbf{Base Loss} & \textbf{MG Loss}    &    \textbf{Speedup}\\
        \hline
        Half-V Cycle (no network adaptation) &   21860.50 & 12270.44  & 0.0050 &  0.0067 & 1.94$\times$ \\
        \hline
        Half-V Cycle (network adaptation) & 36267.75 &  11803.04 &  0.0047  &  0.0052 & 3.07$\times$ \\
        \hline 
    \end{tabular}
    \end{center}
\end{table*}

\subsection{Multigrid Training}
We begin this study by first sampling the set of coefficients $\underline{\omega}$ used for generating the diffusivity maps using \Eqnref{def:poisson-nu-harmonic-expansion}. We sampled a  total of $65536$ coefficients using a quasi-random Sobol sampling methodology. As stated earlier, we use a U-Net architecture for performing all the experiments. The U-Net used has a depth of 3 (i.e., a total of 3 convolution layers and 3 transpose convolution layers). First, a block of convolution and batch normalization is applied. Then, the output is saved for later use using the skip-connection. This intermediate output is then downsampled to a lower resolution for a subsequent block of convolution, batch normalization layers. This process is continued two more times. The upsampling starts where the saved outputs of similar dimensions are concatenated with the upsampling output for creating the skip-connections followed by a  convolution layer. LeakyReLU activation was used for all the intermediate layers. The final layer has a Sigmoid activation. The starting filter size is 16, and we double the number of filters as the depth of the U-Net increases. For all the studies, we use Adam optimizer~\citep{Kingma2015AdamAM} with a learning rate of $1\times10^{-5}$ and the global batch size of 64.

\subsubsection{Multigrid Strategies:}
We begin by studying each multigrid strategy at different resolutions. In \Tableref{Table:strategies}, we provide a detailed study on time taken to achieve convergence and the loss achieved. As our baseline, we perform full training at the highest resolution of the multigrid to quantify the performance. The time and the loss value at convergence for this full training are reported as Base Time and Base Loss. First, we note that all the strategies at all the resolutions converge around the similar loss value compared to the Base Loss. Also, at lower resolutions, the speedup obtained from the multigrid approaches is very marginal, and for the Half-V cycle, it is worse than the Base training time. At the same time, the V cycle has the best computational speedup. 

\begin{figure}[h!]
    \centering
    \includegraphics[width=0.65\linewidth,trim={0in 2in 0in 2in},clip]{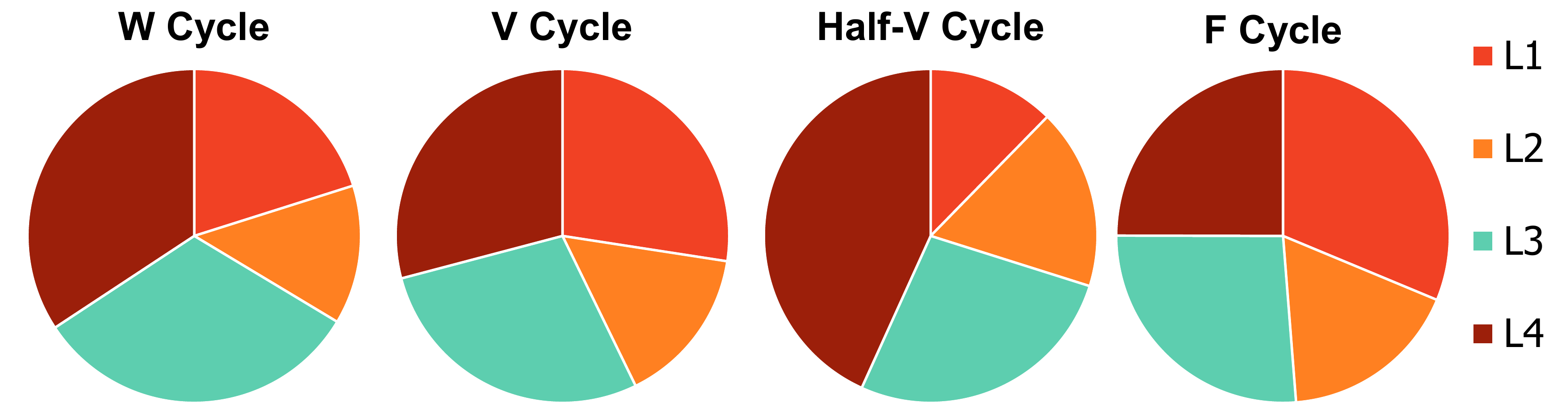}
    \caption{Pie chart for \% time spent on training at different resolutions for each multigrid strategy}
    \label{fig:piechart_timespent}
\end{figure}

The speedup increases with the increase in resolution for each strategy (except for the F cycle, where the increase is marginal). We also see that each strategy has a slightly different trend in speedup with the increase in resolution. To understand this, we plot the $\%$ time spent on each of the levels of resolution in \Figref{fig:piechart_timespent}. With the primary assumption that $\%$ time spent on lower resolutions is better than that on higher resolution (based on \Figref{fig:epochtimes}), we understand that the Half-V cycle is the best. However, at lower resolutions such as $128\times128$, the time taken per epoch on the lower resolution is comparable with the time taken per epoch on higher resolution. This allows for a drastic jump in speedup from $128\time128$ to $512\time512$. At the same time, the speedup for the V cycle increased and then reduced. While the speedup is desired, we want the \mgdiffnet{} to have similar performance accuracy compared to the base network. Consistently in all the resolutions, Half-V and F cycles perform much closer to the Base loss, whereas the V cycle has the maximum deviation from the Base loss. Combined with the fact that the Half-V cycle has a much better speedup than the F cycle, we conclude that the Half-V cycle performs the best among all the strategies for this problem. In the following experiments, we only show results on the Half-V cycle \mgdiffnet{} strategy.

\subsubsection{Architectural Adaptation:}
A direct extension to the proposed multigrid approach is to adaptively add more weights for performing better at higher resolutions. This is particularly interesting when the assumption that the network has infinite learning capacity is relaxed. As soon as this assumption is relaxed, one can question if the network learning at a lower resolution is sufficient for learning at higher resolutions. To evaluate this question, we perform an experiment where we add three additional layers (one convolutional layer and two transpose convolutional layers) and remove one learned transpose convolutional layer after training at each coarse resolution and moving to the finer resolution. The additional layers added are again initialized with random weights. However, we observe that within 20-30 mini-batches of update, the loss (which is expected to rise due to the random weights) drops down. \Tableref{Table:adaptation}, shows comparisons between with and without adaptation. Note that the base time and base loss for the case with architectural adaptation accounts for the final network architecture and an experiment to run full training on that final network architecture. We note that there is a marginal improvement in the loss at the same time; we show that there is a $3\times$ improvement in training time for a very deep U-Net architecture. This ties into the theme of correlations between U-Net architecture and multigrid methods mentioned in \citet{alt2021translating}.

\subsubsection{Scaling to 3D:}
\begin{figure}[h!]
    \centering
    \includegraphics[width=0.55\linewidth,trim={0in 0in 0in 0in},clip]{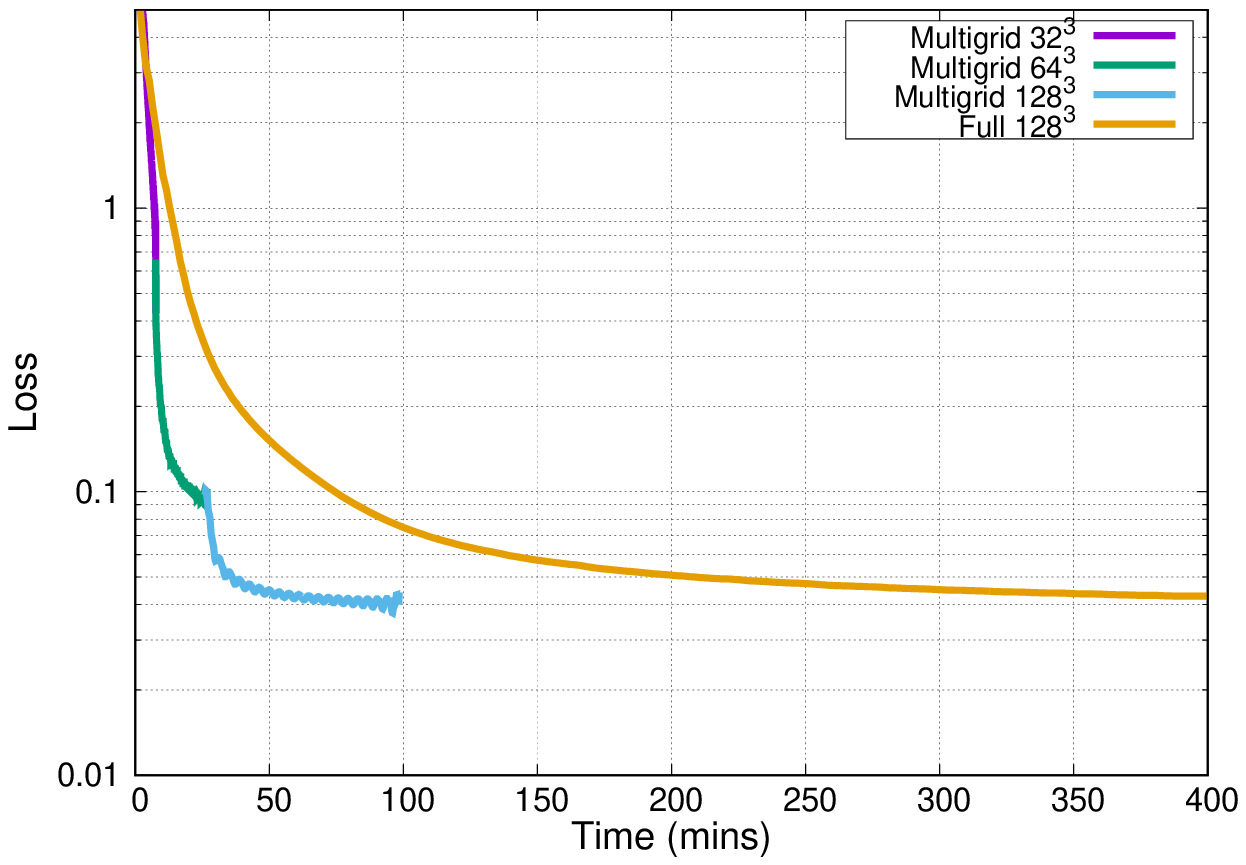}
    \caption{Comparison of performance of base training and multigrid training for $128\times128\times128$ resolution. The multigrid strategy used here is the Half-V cycle.}
    \label{fig:perf_plot_3d_128}
\end{figure}

With both the architectural adaptation and Half-V cycle, in 2D spatial domain with a resolution of $512\times512$, we get a speedup of $3\times$ over the baseline training approach at full resolution. We now extend this framework to solve PDEs at higher resolutions in 3D. In \Tableref{Table:strategies}, we show the result of a $128\times128\times128$ resolution of field outputs. we see that our network performs similarly to the base network while achieving a total speedup of $6\times$. We also show the loss performance plot of our multigrid approach in comparison with full training at the same resolution in \Figref{fig:perf_plot_3d_128}. We see that the losses are first reduced in the lower resolutions and then further reduced at a finer resolution (as anticipated in a multigrid solver).

\subsection{Scaling to Significantly Higher Resolutions} \label{sec:scaling}
In what follows, we demonstrate the ability to train 3D \mgdiffnet{} on much higher resolutions by scaling out on GPU and CPU clusters with hundreds to thousands of cores. We show that we can achieve excellent speedups on both cloud and bare-metal infrastructures.

\subsubsection{Scaling on a GPU Cluster:} \label{sec:scaling_gpu}
The first set of experiments were performed on a GPU cluster of NDv2-series VMs on Microsoft Azure, each containing 8 NVIDIA Tesla V100 GPUs with 32GB of memory per device. The input dataset consisted of 1024 parametric diffusivity maps of size $256\times 256\times 256$, as described by \Eqnref{def:poisson-nu-harmonic-expansion}. The training was performed on clusters with as many as 64 nodes (512 GPUs), using 8 devices per node for $p \ge 8$ processes (for $p < 8$, certain GPUs were left idle). The local mini-batch size was fixed at 2 since each sample required $\sim$14GB during training, and the SGD-based Adam optimizer~\citep{Kingma2015AdamAM} (with a learning rate of $1\times 10^{-4}$) was used.

\begin{figure}[t!]
    \centering
    \includegraphics[width=0.6\linewidth,trim={0in 0in 0in 0in},clip]{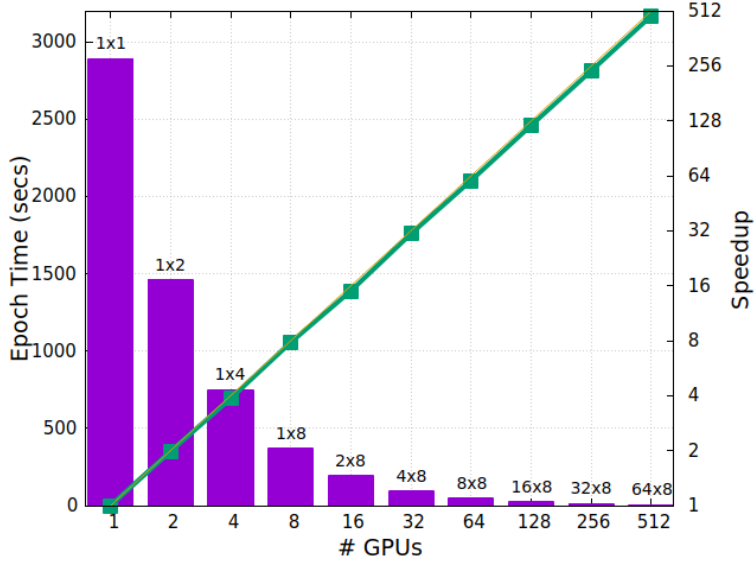}
    \caption{Strong scaling results for training a 3D DiffNet using our distributed deep-learning framework at $256\times256\times256$ resolution on a cluster of NVIDIA Tesla V100 GPUs on cloud. The labels above the bars indicate the number of nodes and the number of GPUs per node.}
    \label{fig:scaling_gpu}
\end{figure}

\Figref{fig:scaling_gpu} shows the wall-clock time per epoch, as well as the corresponding speedup. It demonstrates the ability of our distributed deep-learning solution to scale virtually linearly to 512 GPUs, reducing the runtime per epoch from 48 mins to only 6 secs (a speedup of 480$\times$). Inference time (i.e., full-field prediction time) on a single GPU at this resolution was half a second.

\begin{figure}[t!]
    \centering
    \includegraphics[width=0.6\linewidth,trim={0in 0in 0in 0in},clip]{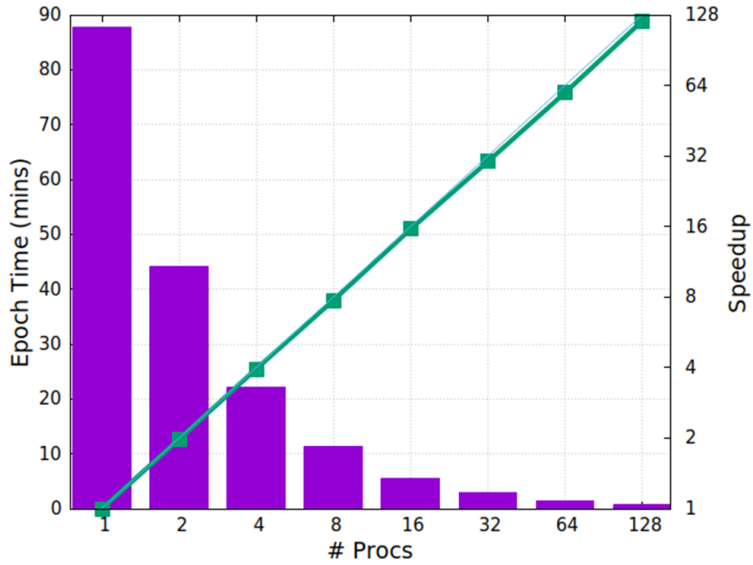}
    \caption{Strong scaling results for training a 3D DiffNet using our distributed deep-learning framework at $512\times512\times512$ resolution on a cluster of AMD EPYC-7742 bare-metal nodes (with 1 process per node).}
    \label{fig:scaling_cpu}
\end{figure}

\begin{table}[!t]
	\centering
	\small
	\newcommand\T{\rule{0pt}{2.7ex}}
	\newcommand\B{\rule[-1.3ex]{0pt}{0pt}}
	\newcommand{\tabincell}[2]{\begin{tabular}{@{}#1@{}}#2\end{tabular}}
	\tymin=.1in
	\tymax=2.5in 
    \caption{Visualization of \mgdiffnet{} predictions with different multigrid strategies. The input $ \underline{\omega} = (0.3105,  1.5386,  0.0932, -1.2442) $}.
    \label{Tab:DiffNetStrategies}
	\begin{tabular}{lcc}
        \textbf{Strategy} & \textbf{$u_{\mgdiffnet{}}$} & \textbf{$u_{\mgdiffnet{}} - u_{FEM}$}\\
        \hline
        V Cycle & \tabincell{c}{\includegraphics[width=0.25\linewidth,trim={0 0 0 0},clip]{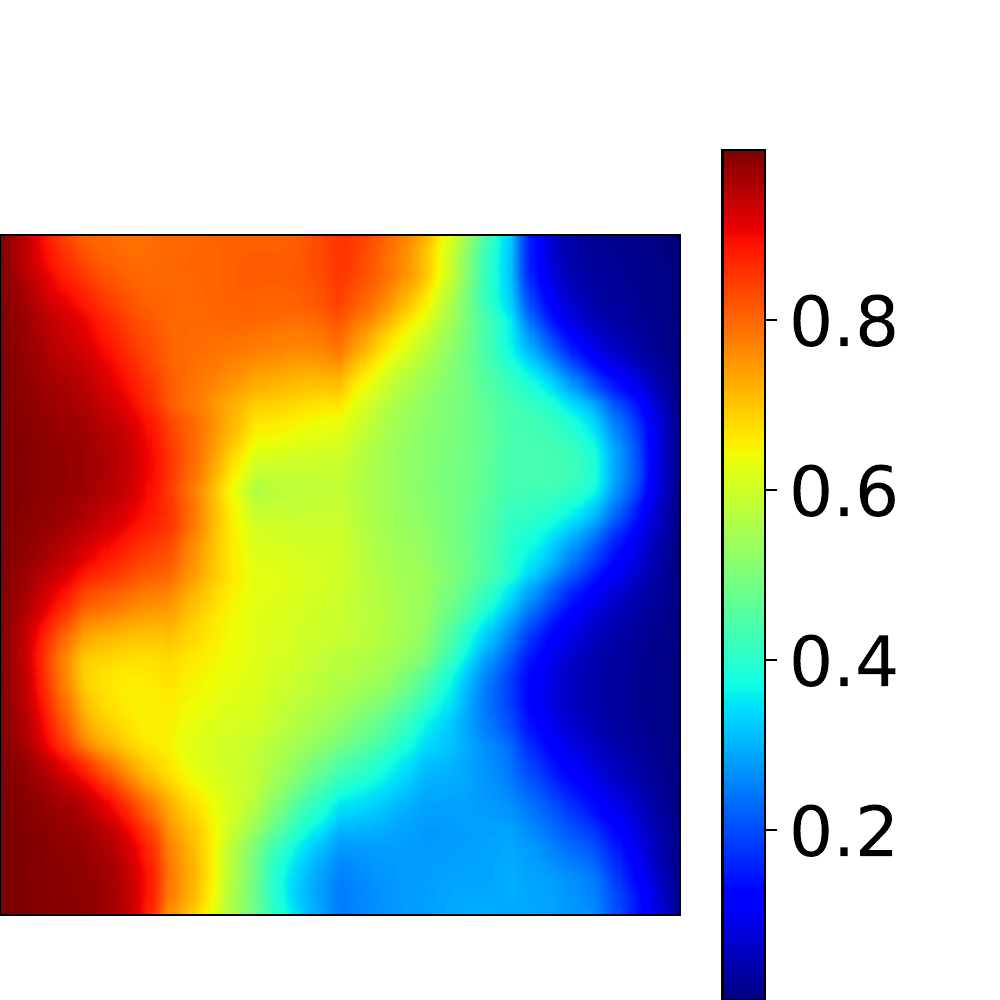}}&	\tabincell{c}{\includegraphics[width=0.25\linewidth,trim={0 0 0 0},clip]{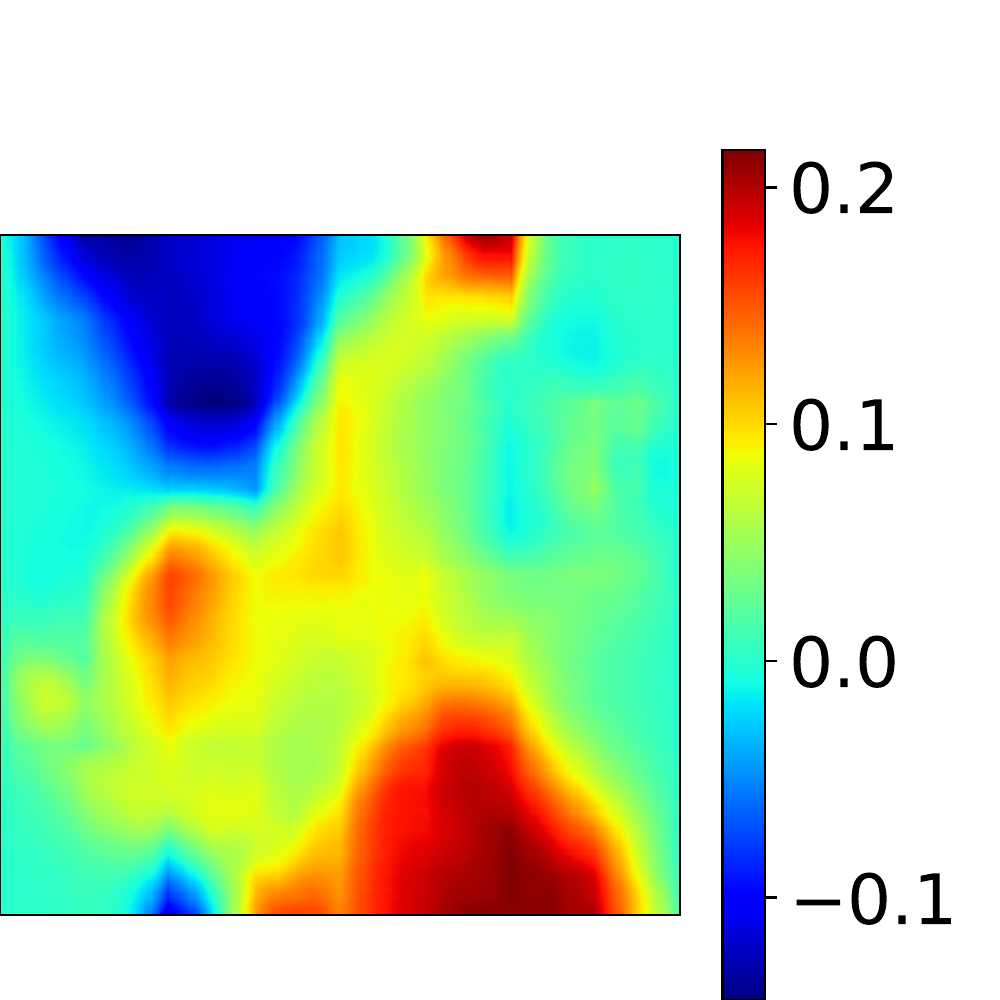}}\\
        \hline
        W Cycle & \tabincell{c}{\includegraphics[width=0.25\linewidth,trim={0 0 0 0},clip]{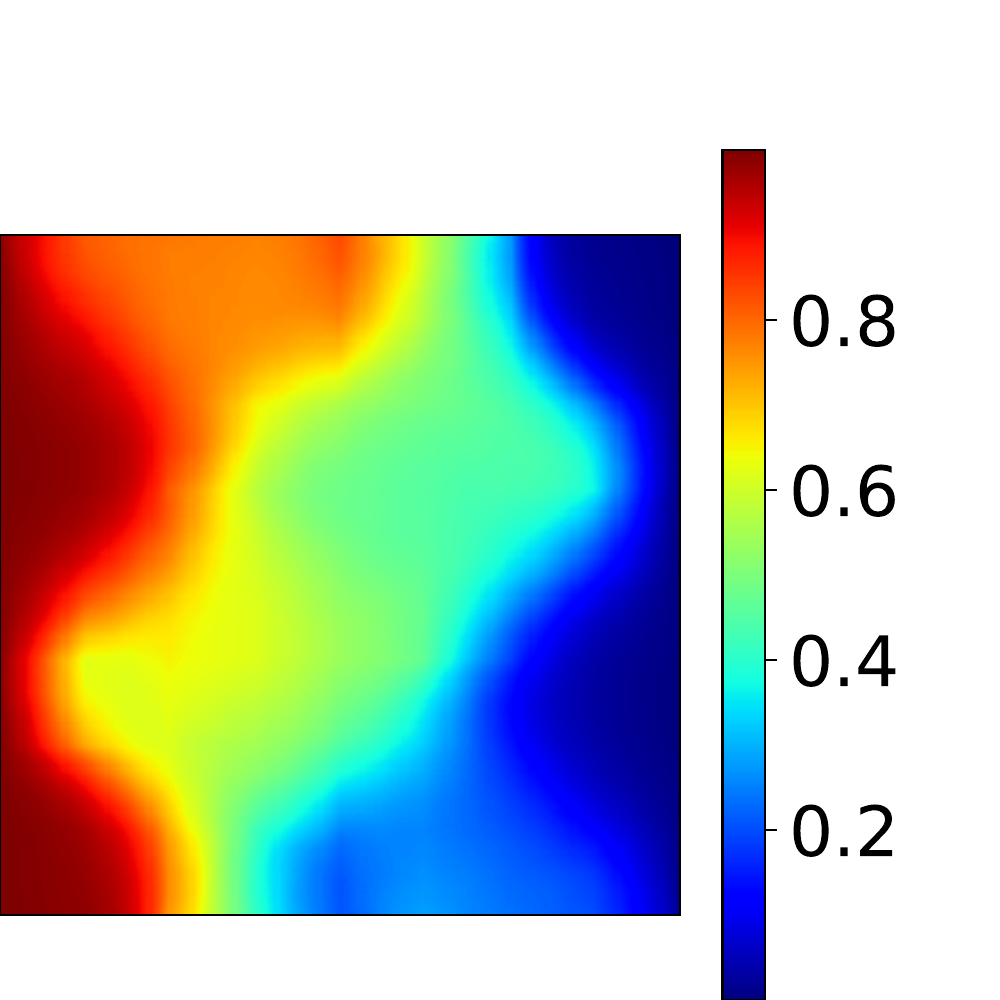}}&		\tabincell{c}{\includegraphics[width=0.25\linewidth,trim={0 0 0 0},clip]{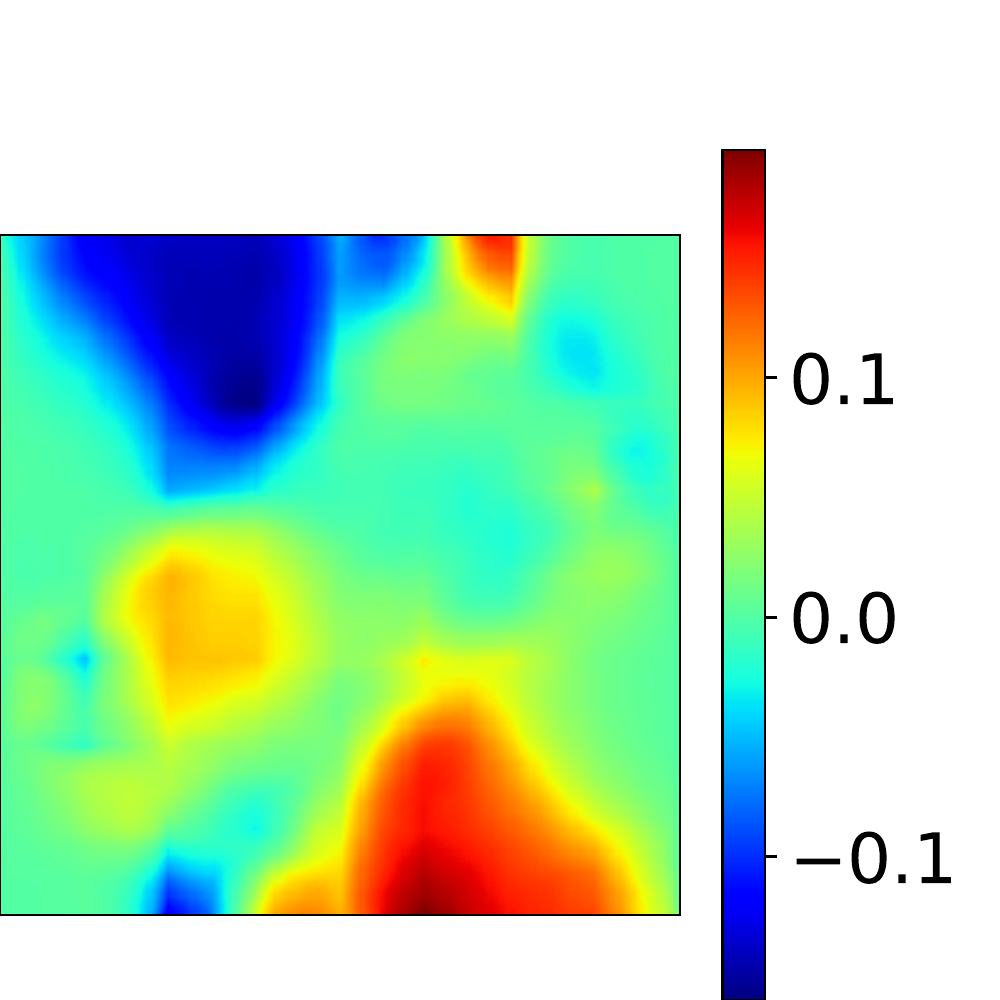}}\\
        \hline
        F Cycle & \tabincell{c}{\includegraphics[width=0.25\linewidth,trim={0 0 0 0},clip]{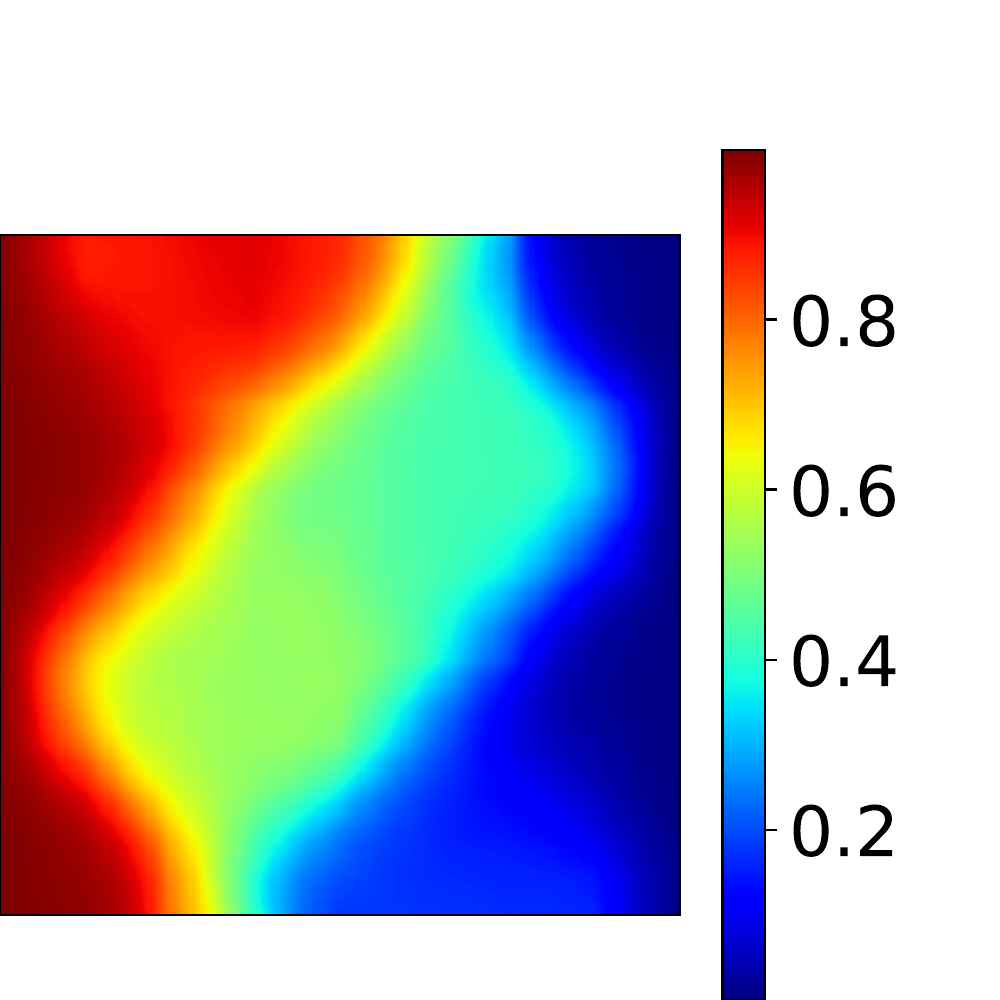}}&		\tabincell{c}{\includegraphics[width=0.25\linewidth,trim={0 0 0 0},clip]{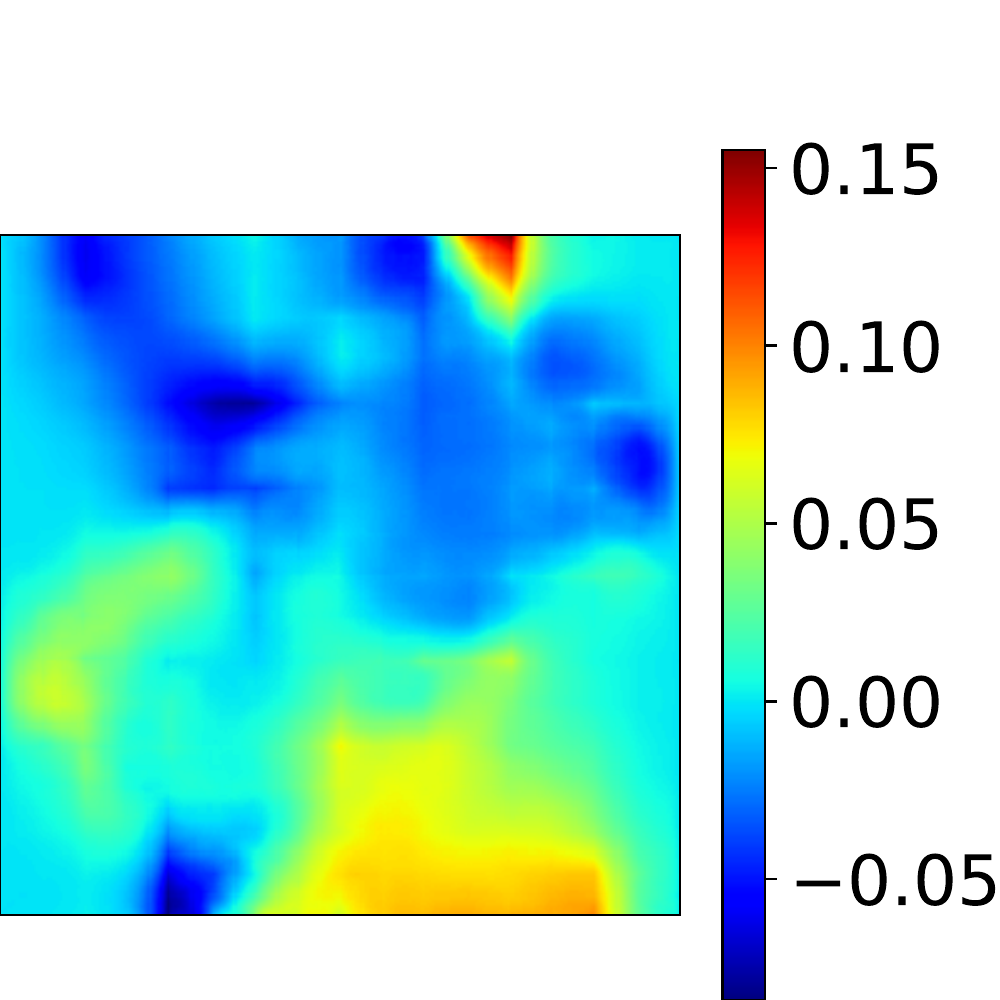}}\\
        \hline
        Half-V Cycle & \tabincell{c}{\includegraphics[width=0.25\linewidth,trim={0 0 0 0},clip]{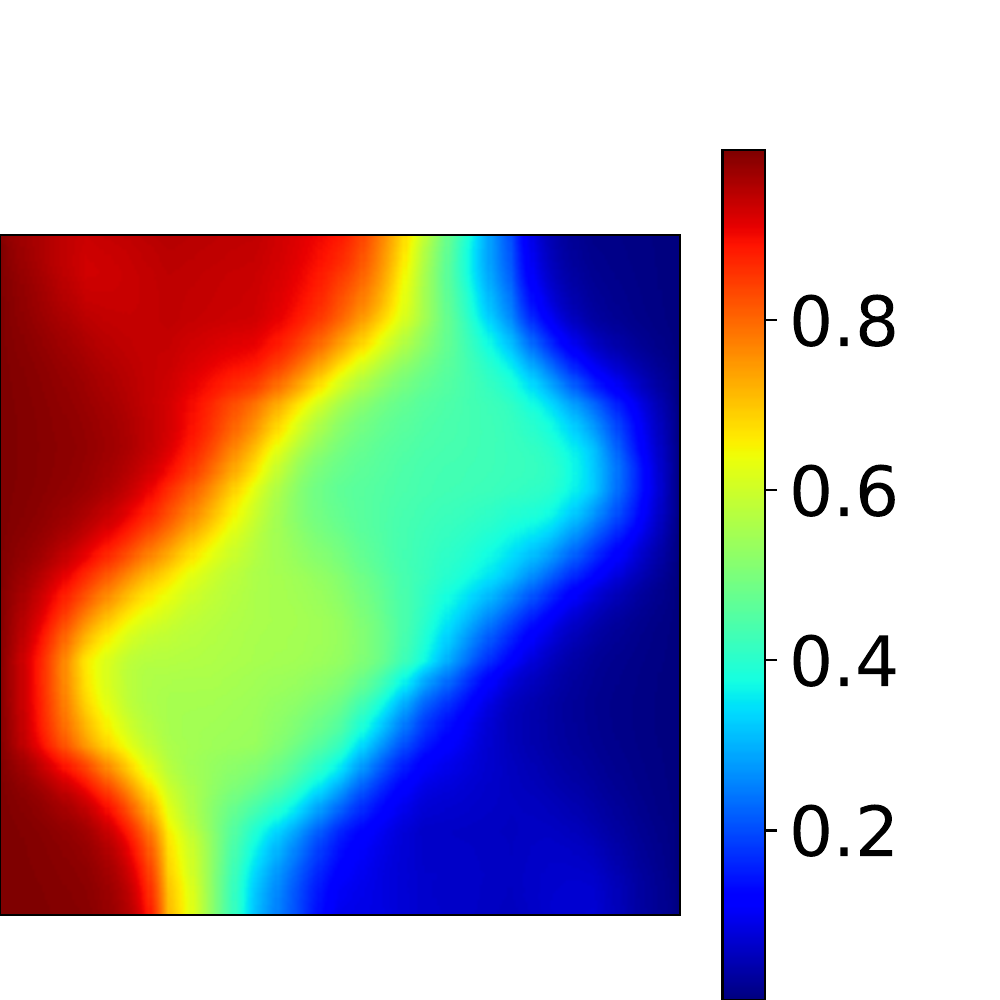}}&		\tabincell{c}{\includegraphics[width=0.25\linewidth,trim={0 0 0 0},clip]{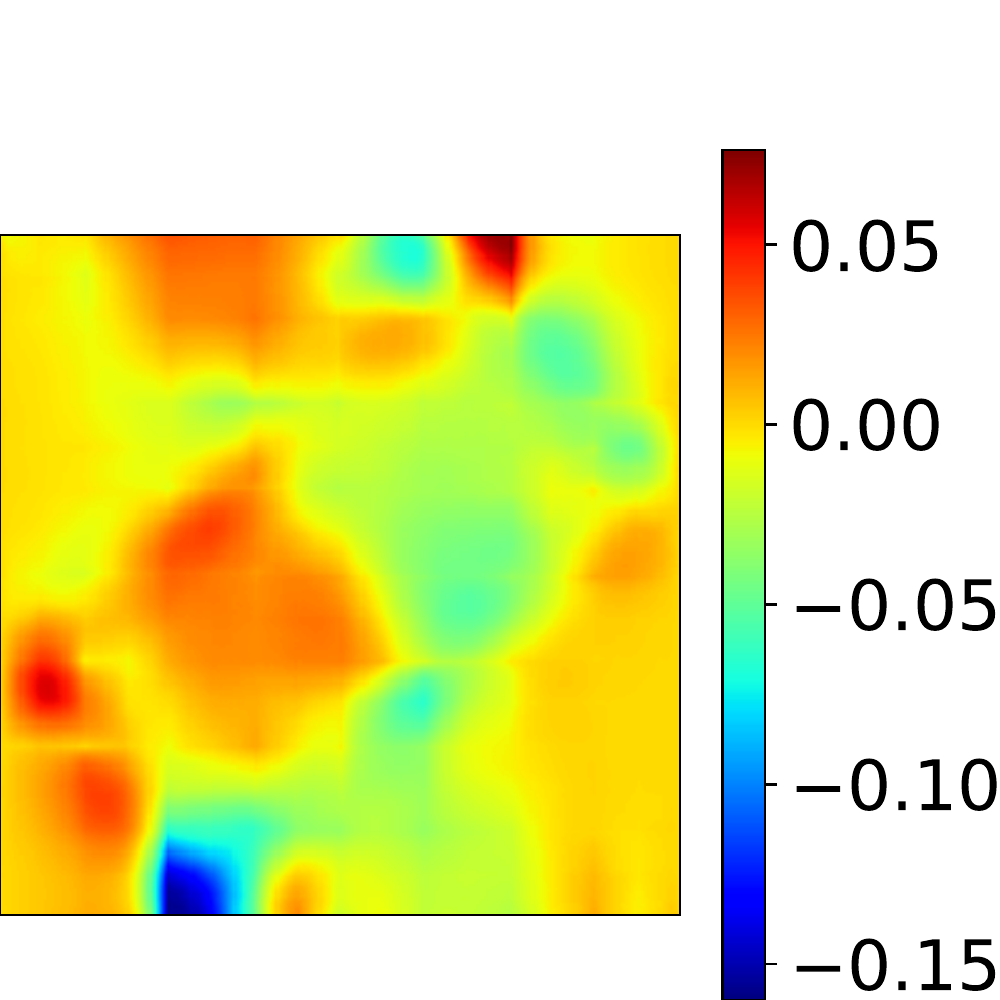}}\\
        \hline
    \end{tabular}
\end{table}

\begin{table*}[!t]
	\centering
	\small
	\newcommand\T{\rule{0pt}{2.7ex}}
	\newcommand\B{\rule[-1.3ex]{0pt}{0pt}}
	\newcommand{\tabincell}[2]{\begin{tabular}{@{}#1@{}}#2\end{tabular}}
	\tymin=.1in
	\tymax=2.5in 
    \caption{Visualization of \mgdiffnet{} predictions and comparison with traditional FEM solutions for 2 anecdotal values of $\underline{\omega}$. }
    \label{Tab:DiffNetPrediction2D-Part1}
	\begin{tabular}{cccc}
    $\nu$	& $u_{\mgdiffnet{}}$ & $u_{FEM}$ & $u_{\mgdiffnet{}} - u_{FEM}$\\
		\hline 
		\tabincell{c}{\includegraphics[width=0.2\linewidth,trim={0 0 0 0},clip]{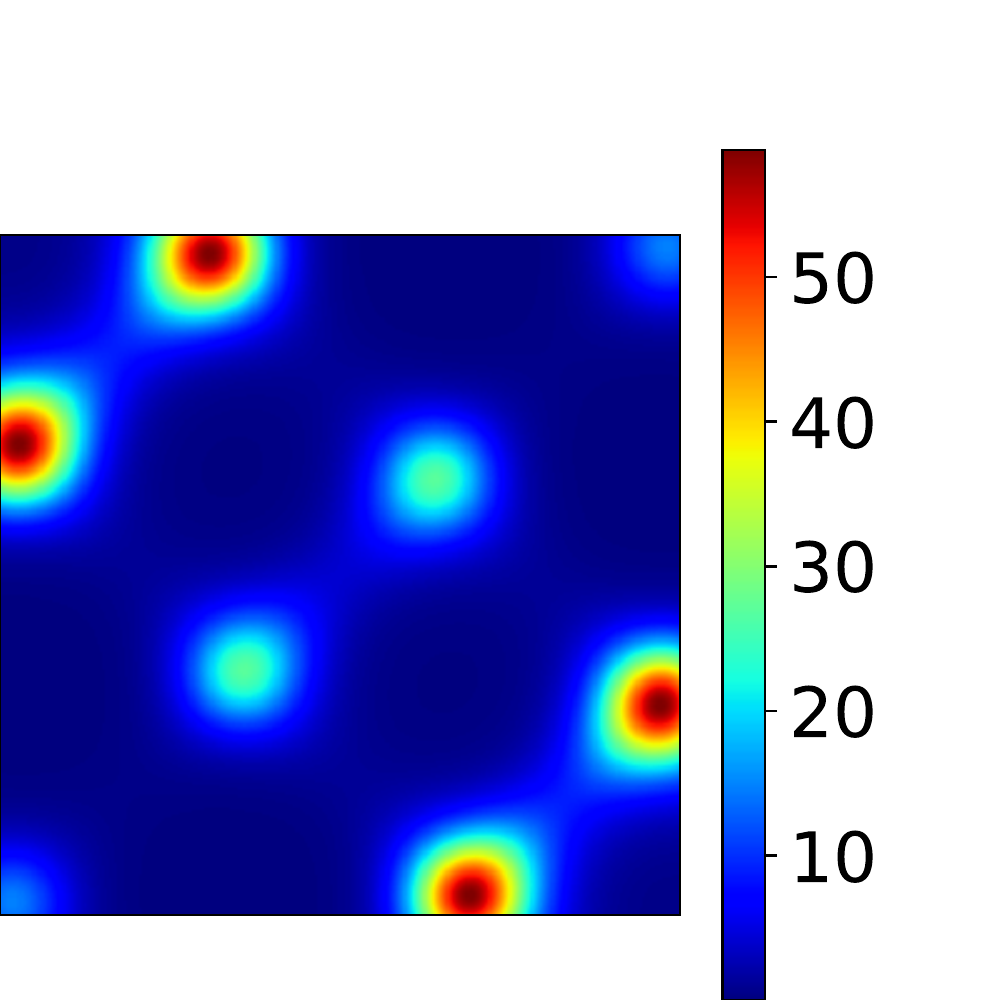}}&
		\tabincell{c}{\includegraphics[width=0.2\linewidth,trim={0 0 0 0},clip]{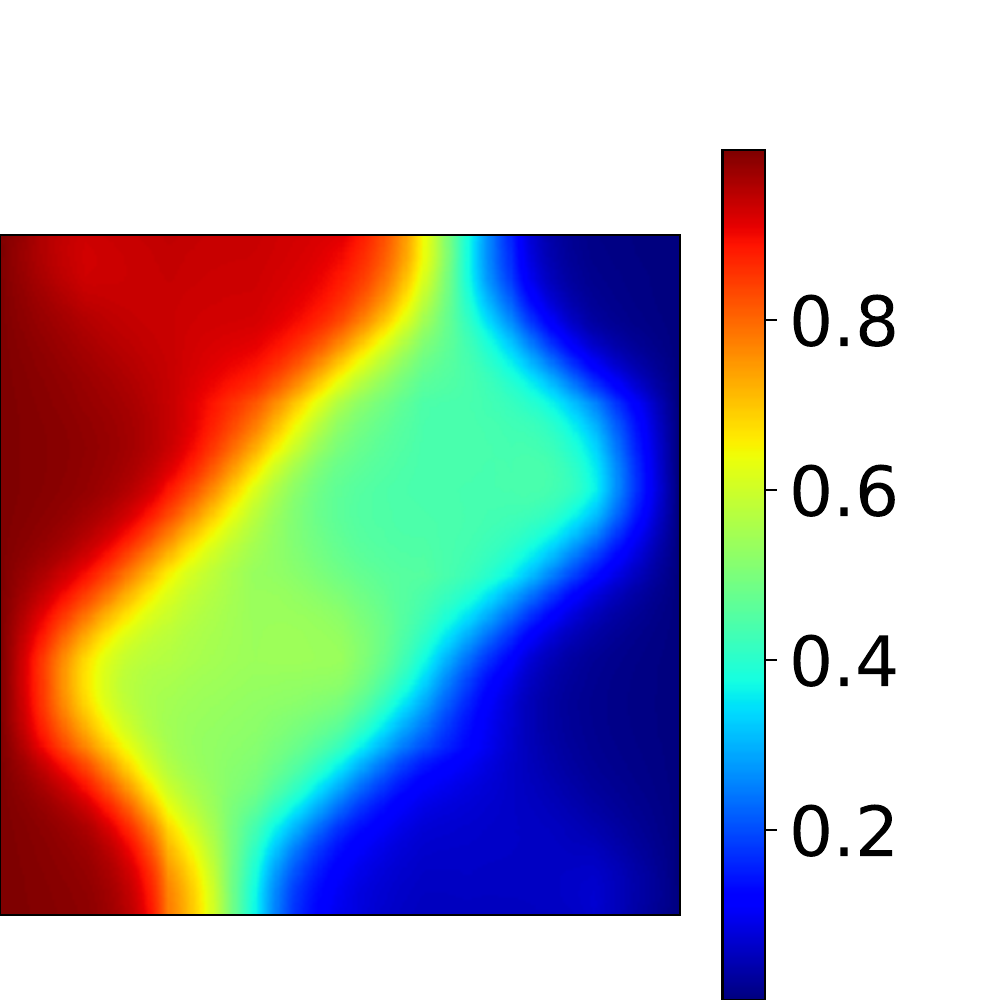}}&
		\tabincell{c}{\includegraphics[width=0.2\linewidth,trim={0 0 0 0},clip]{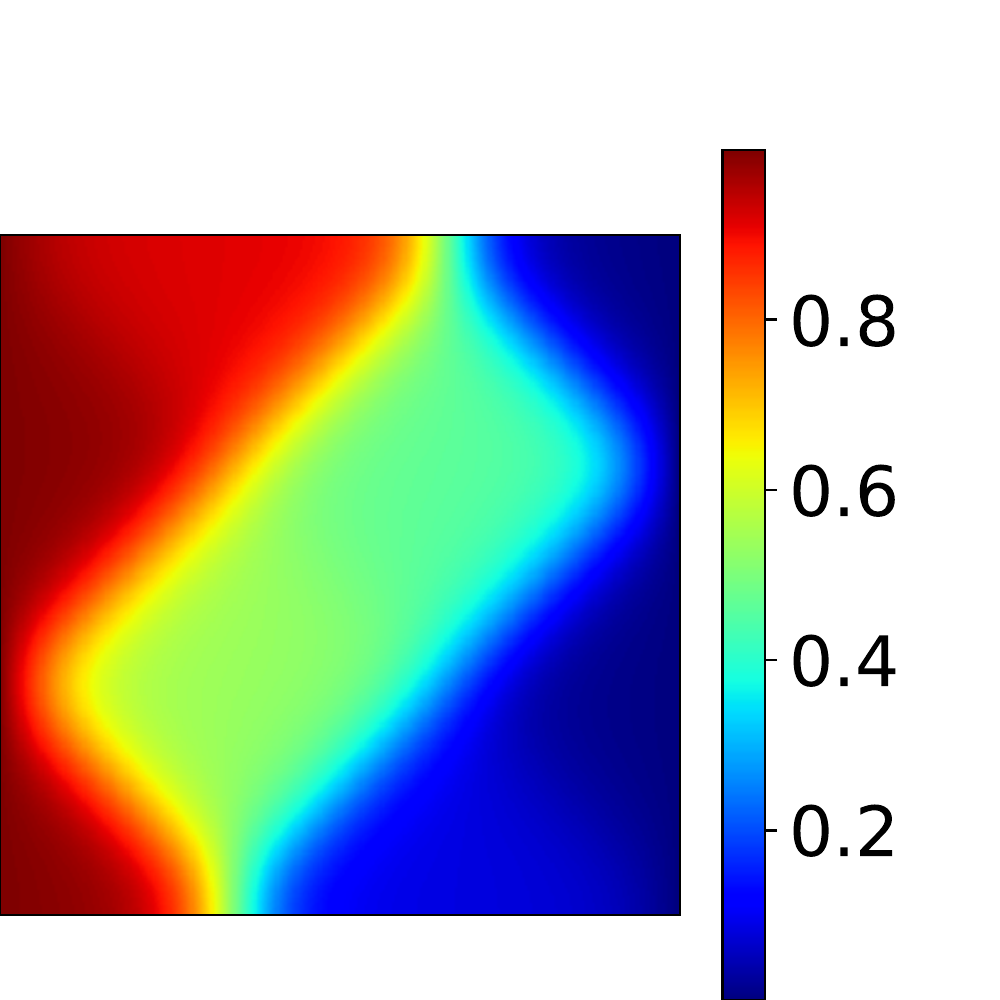}}&		\tabincell{c}{\includegraphics[width=0.2\linewidth,trim={0 0 0 0},clip]{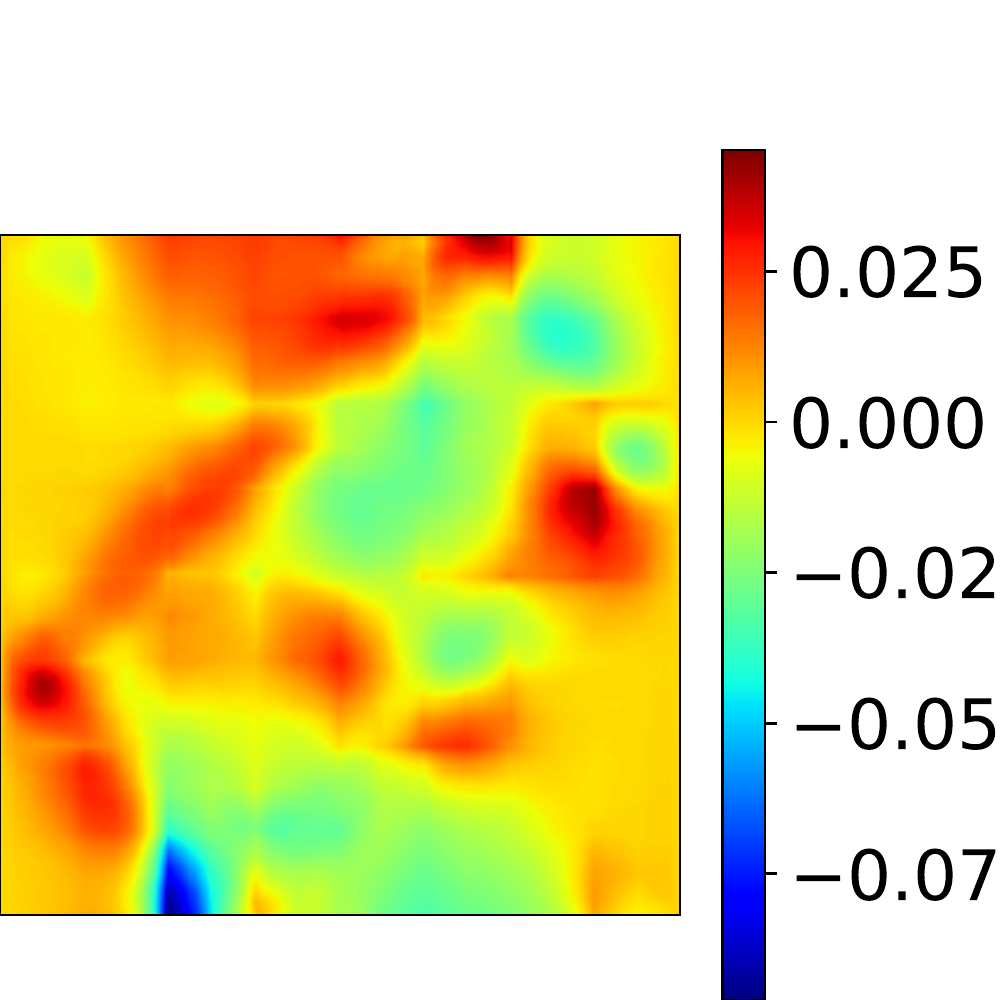}}\\
		&  \multicolumn{2}{c}{$ \underline{\omega} = (0.6681,  1.5354,  0.7644, -2.9709) $} &\\
		\hline 
		\tabincell{c}{\includegraphics[width=0.2\linewidth,trim={0 0 0 0},clip]{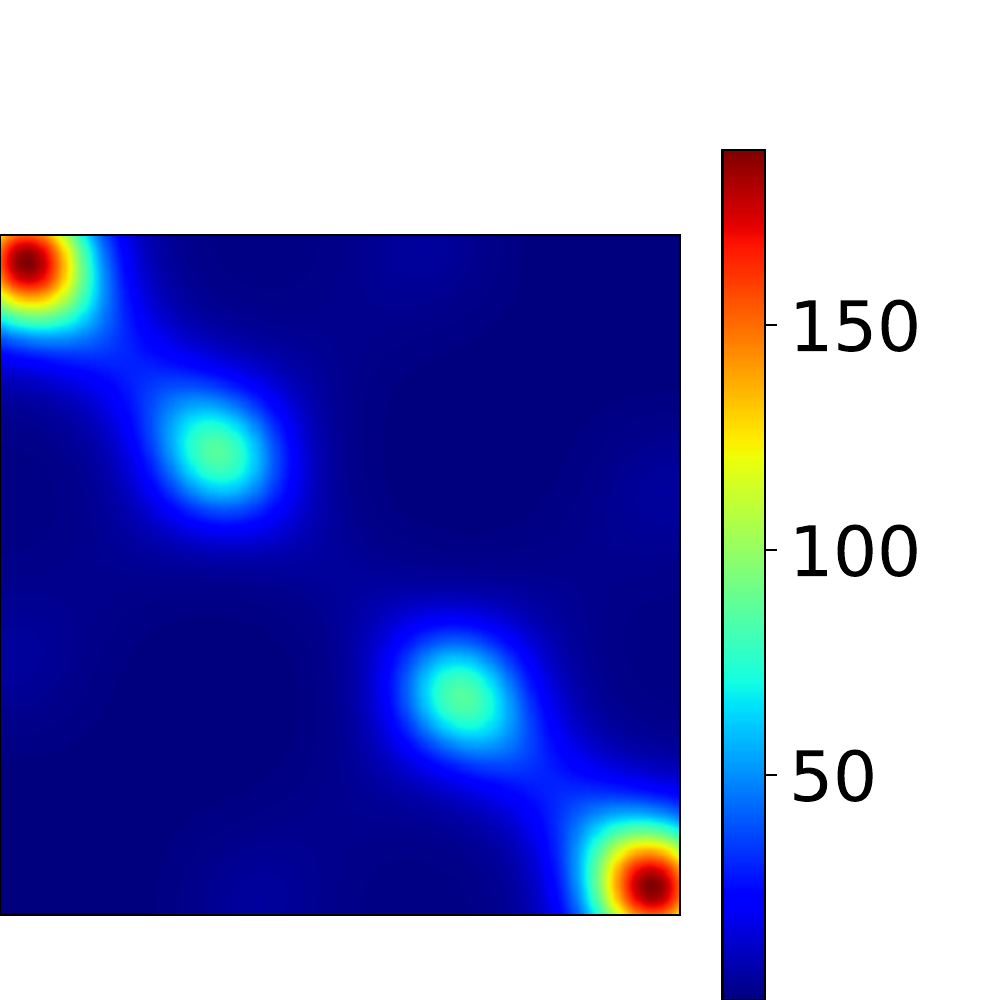}}&
		\tabincell{c}{\includegraphics[width=0.2\linewidth,trim={0 0 0 0},clip]{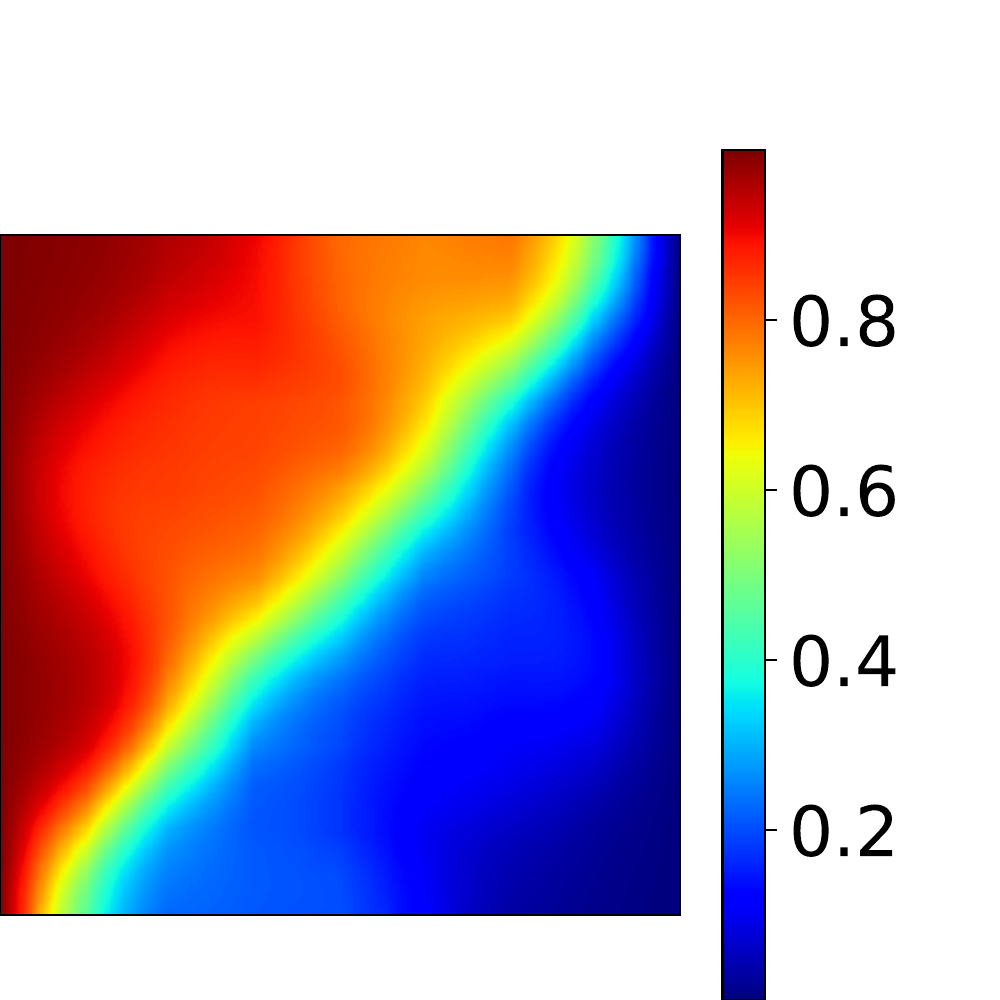}}&
		\tabincell{c}{\includegraphics[width=0.2\linewidth,trim={0 0 0 0},clip]{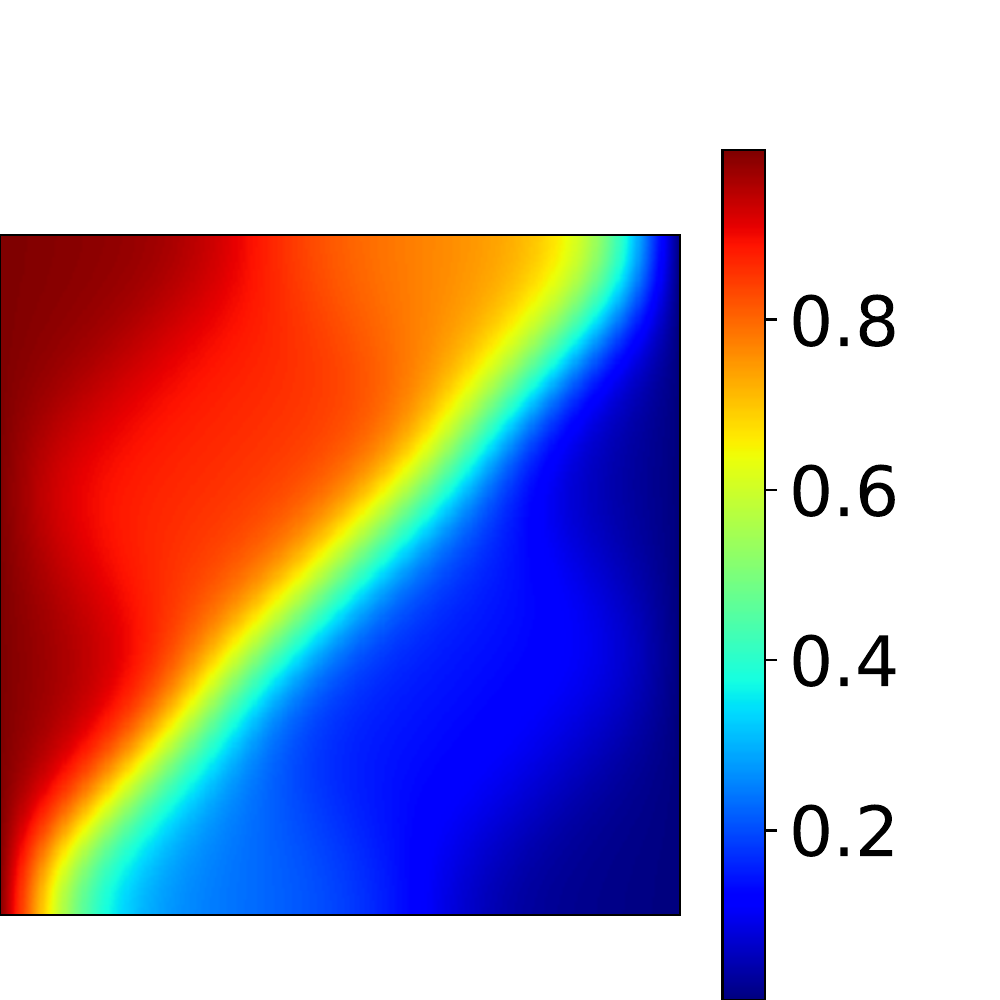}}&		\tabincell{c}{\includegraphics[width=0.2\linewidth,trim={0 0 0 0},clip]{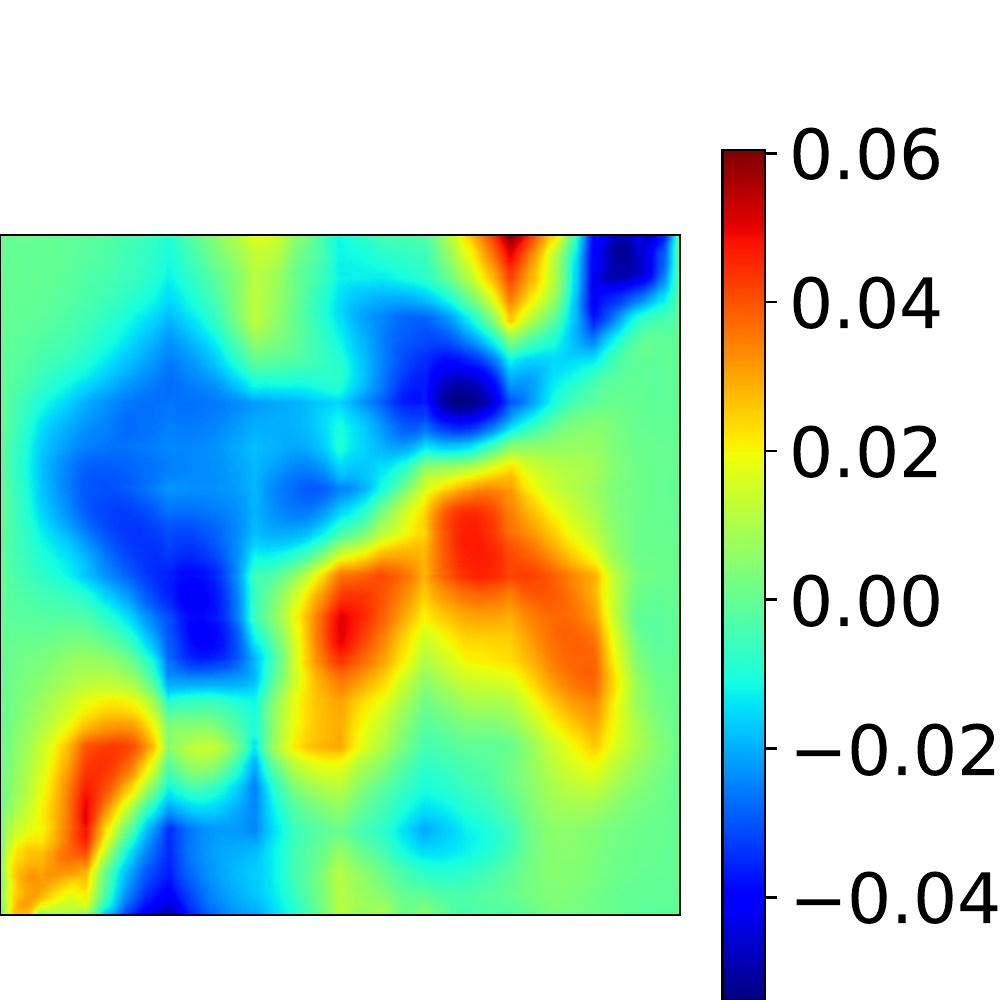}}\\
		&  \multicolumn{2}{c}{$ \underline{\omega} = (1.3821,  2.5508,  0.1750,  2.1269) $} &\\
		\hline 
    \end{tabular}
\end{table*}

\begin{table*}[!t]
	\centering
	\small
	\newcommand\T{\rule{0pt}{2.7ex}}
	\newcommand\B{\rule[-1.3ex]{0pt}{0pt}}
	\newcommand{\tabincell}[2]{\begin{tabular}{@{}#1@{}}#2\end{tabular}}
	\tymin=.1in
	\tymax=2.5in 
    \caption{Visualization of \mgdiffnet{} predictions and comparison with traditional FEM solutions for $\underline{\omega} = (0.3105,  1.5386,  0.0932, -1.2442)$. }
    \label{Tab:DiffNetPrediction3D}
	\begin{tabular}{ccc}
    $\nu$	& $u_{\mgdiffnet{}}$ & $u_{FEM}$\\
		\tabincell{c}{\includegraphics[width=0.3\linewidth,trim={0 0 0 0},clip]{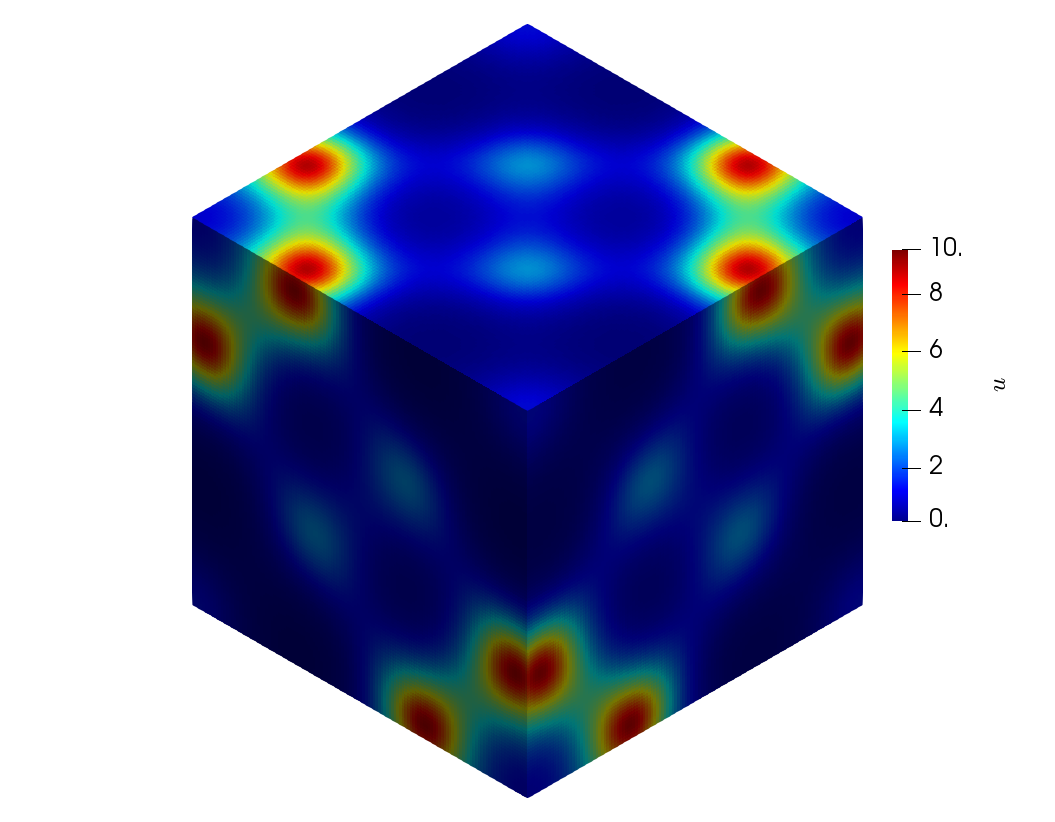}}&
		\tabincell{c}{\includegraphics[width=0.3\linewidth,trim={0 0 0 0},clip]{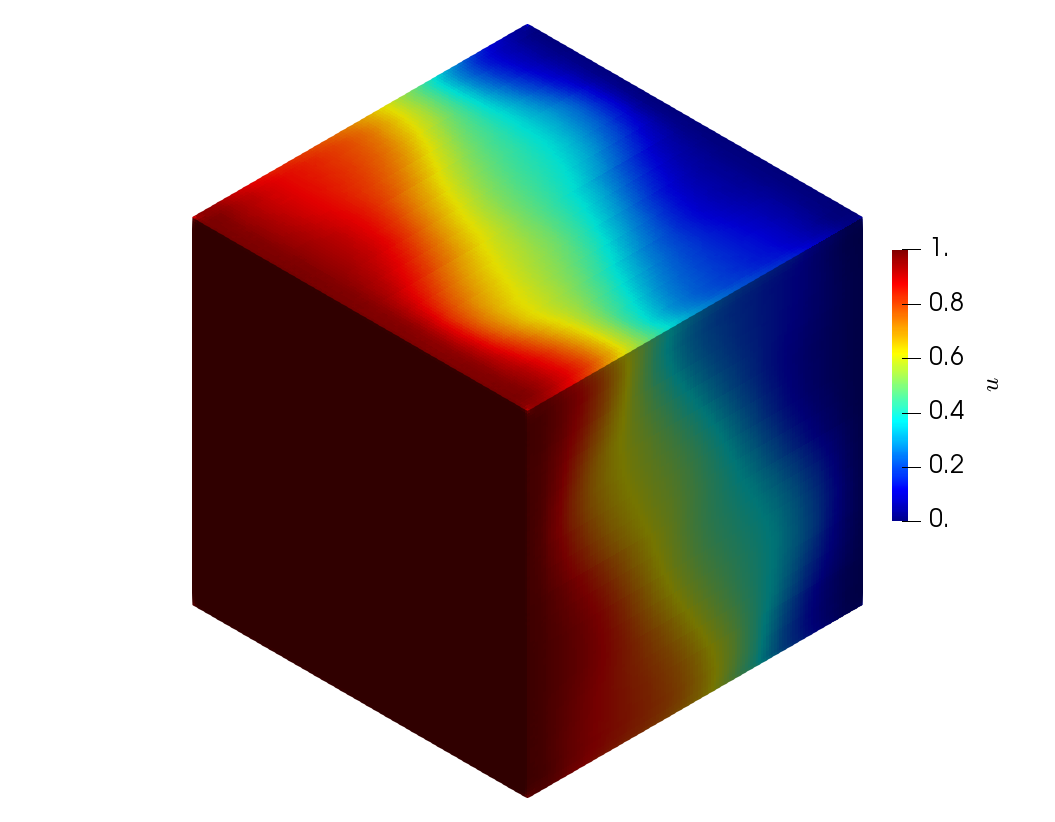}}&
		\tabincell{c}{\includegraphics[width=0.3\linewidth,trim={0 0 0 0},clip]{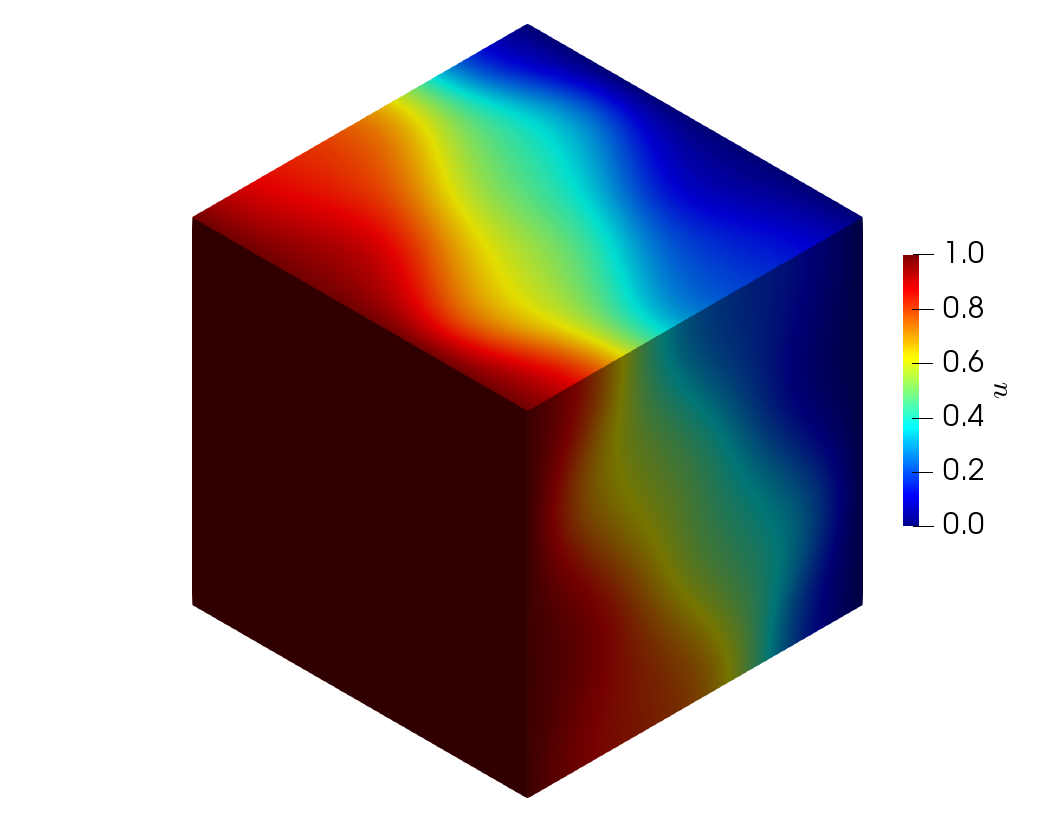}}\\
    \end{tabular}
\end{table*}

\subsubsection{Scaling on a CPU Cluster for Significantly High Resolutions:} \label{sec:scaling_cpu}
Despite achieving excellent speedups, training on GPUs is still limited by their relatively small available memory per device, which caps the maximum size of the training volumes at $256\times 256\times 256$. To demonstrate the ability of our software to solve problems at even higher resolutions, we trained DiffNet with diffusivity maps of size $512\times 512\times 512$ on a cluster of AMD EPYC-7742 CPU nodes, each with 128 cores and 256GB total RAM. \Figref{fig:scaling_cpu} shows epoch times and speedups obtained on clusters with up to 128 nodes, with one MPI process per node (using all 128 CPU cores) and two samples per local batch. Once again, scalability is very strong, up to 128 nodes. The peak memory utilization per node was 230GB, which would have been unfeasible on a cluster of GPUs. The full-field prediction time on the same machine type was of 20 seconds.

\subsection{Comparison with Traditional FEM}
We also provide some visualizations and comparisons with traditional FEM simulations for the same parameters. \Tableref{Tab:DiffNetStrategies} shows the visualization of the predictions from the multigrid trained network for $512\times512$. We see the \mgdiffnet{} predicts the solution field accurately. We also compare the results obtained by different multigrid strategies to confirm that the Half-V cycle predictions are the best among all the strategies. We also show visualization of a few anecdotal solution fields produced using \mgdiffnet{} in 2D (\Tableref{Tab:DiffNetPrediction2D-Part1}) and 3D (\Tableref{Tab:DiffNetPrediction3D}). Another important comparison is the time taken for inference compared to the time taken for performing one finite element solve. While the FEM simulation takes about 5 minutes for $128\times128\times128$ resolution, the \mgdiffnet{} inference takes less than 30 seconds. Since the solutions are valid for a range of PDE parameters, our framework's impact on reducing the computational time while performing inverse design will be much higher. We also note that there is no need for any data annotation in this framework.

\section{Conclusion and Future Work}
In this work, we propose a distributed multigrid neural solver for solving PDEs at large spatial dimensions with efficient use of computational resources. To this end, we contribute a numerical multigrid-inspired training scheme for fully convolutional neural networks and further implement a distributed data-parallel training strategy to train networks up to a resolution of $512\times512\times512$ ($\approx 134M$ voxels). Our multigrid-based training results show a 6X speedup over the baseline full training at higher resolutions with negligible loss in performance. Further, our method scales almost linearly with minimal communication costs in a distributed environment over both CPU and GPU clusters. This approach opens up the efficient training of parametric PDEs for use in Scientific ML applications. Additionally, this approach can be naturally applied to a variety of high-resolution image-to-image translation tasks. 

\pagebreak 

\noindent There are several avenues of future work that follow:
\begin{itemize}[left=0pt,topsep=0.0in]
    \item Scaling beyond megavoxels to gigavoxels (which we hope to accomplish by the time of review of this paper).
    \item Extending our approach to allow \textit{model-parallel} distributed deep learning.
    \item Elucidating the mathematical connections between the multigrid approach with stability and convergence of the training.
    \item  Deploying this neural PDE Poisson solver for applications in topology optimization, flow through porous media, and thermal transport in composites--all of which are defined by \Eqnref{eq:poisson-pde}.
    \item Deploying this framework to other PDE's where having high-resolution outputs is critical for control (via model predictive control approaches).
\end{itemize}
We envision such bidirectional linkages between numerical linear algebra and scalable solutions of neural networks to significantly accelerate scientific computing workflows. 


\clearpage
\newpage
\bibliographystyle{IEEEtranSN}
\bibliography{refs}

\clearpage
\newpage
\appendix
\section*{Appendix}
\subsection*{Summary of Reported Experiments}
We performed the experiments (all experiments are described in the ``Results and Discussions" section of the paper):
\begin{enumerate} 
\item Comparison of strategies - these were done on Azure cloud platform.
\item Scaling studies were performed for training \mgdiffnet{} of $256\times256\times256$ and lower were performing on Azure cloud platform and studies above $256\times256\times256$ were performed on PSC Bridges2.
\item Solving the PDE using FEM for comparison with \mgdiffnet{} results was done on PSC Bridges2 using 1 Regular Memory node.
\end{enumerate}
Modules loaded on Bridges2 for \mgdiffnet{} experiments:
\begin{verbatim}
1) cmake/3.16.1  
2) gcc/10.2.0
3) openmpi/4.0.5-gcc10.2.0
\end{verbatim}

\subsection*{Libraries Dependencies}
The following dependencies are required to compile the code:
\begin{itemize}
	\item C/C++ compilers with C++11 standards and OpenMP support
	\item MPI implementation (e.g. openmpi, mvapich2 )
	\item Petsc 3.8 or higher
	\item ZLib compression library (used to write \texttt{.vtu} files in binary format with compression enabled)
	\item  MKL / LAPACK library
	\item CMake 2.8 or higher version
	\item OpenCV 3.4.2
\end{itemize}


\subsection*{Computing Configuration}
Relevant computational hardware details are provided here:

\begin{table}[!b]
    \setlength\extrarowheight{3pt}
	\newcommand{\tabincell}[2]{\begin{tabular}{@{}#1@{}}#2\end{tabular}}
    \centering
    \caption{Functional specifications of Microsoft Azure and Bridges2 infrastructures used in our experiments.}
    \label{tab:azure_vs_bridges_specs}
    \begin{tabular}{|c|c|c|}
        \hline
        {\bf Specification} & {\bf Microsoft Azure}  & {\bf Bridges2} \\
        \hline
        Type & Virtual Machine & Bare-Metal \\
        \hline
        CPU & \tabincell{c}{Intel Xeon\\Platinum 8168} & AMD EPYC 7742 \\
        \hline
        CPU cores & 40 & 128 \\
        \hline
        Memory (GB)& 672& 256\\
        \hline
        GPU & Tesla V100 & - \\
        \hline
        GPU Memory (GB)  & 32 & - \\
        \hline
        No. of GPUs & 8 & - \\
        \hline
        Interconnect & EDR Infiniband & HDR Infiniband \\
        \hline
        Bandwidth & 100 Gb/sec & 200 Gb/sec \\
        \hline
        Topology & Fat tree & Fat tree \\
        \hline
    \end{tabular}
\end{table}

\pagebreak









\section{Additional Examples}
We provide few anecdotal evaluations of \mgdiffnet{} for different $ \underline{\omega}$ values sampled in the same range as the training samples. We also provide  the comparison with FEM solutions for the same in \Tabref{Tab:DiffNetPrediction2D-Part2}.

\begin{table*}[h!]
	\centering
	\small
    \setlength\extrarowheight{5pt}
	\newcommand\T{\rule{0pt}{2.7ex}}
	\newcommand\B{\rule[-1.3ex]{0pt}{0pt}}
	\newcommand{\tabincell}[2]{\begin{tabular}{@{}#1@{}}#2\end{tabular}}
	\tymin=.1in
	\tymax=2.5in 
    \caption{Visualization of \mgdiffnet{} predictions and comparison with traditional FEM solutions for 5 anecdotal values of $\underline{\omega}$. }
    \label{Tab:DiffNetPrediction2D-Part2}
	\begin{tabular}{cccc}
    $\nu$	& $u_{\mgdiffnet{}}$ & $u_{FEM}$ & $u_{\mgdiffnet{}} - u_{FEM}$\\
        \hline
		\tabincell{c}{\includegraphics[width=0.2\linewidth,trim={0 0 0 0},clip]{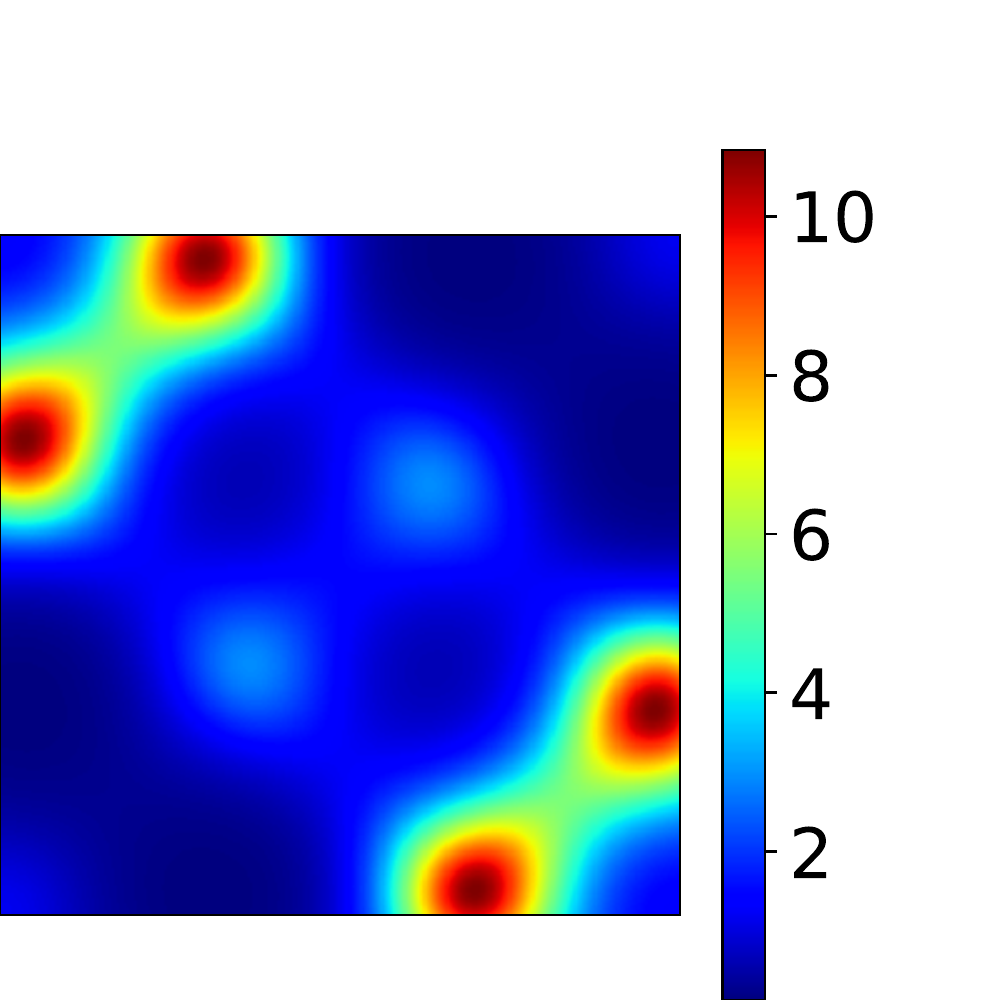}}&
		\tabincell{c}{\includegraphics[width=0.2\linewidth,trim={0 0 0 0},clip]{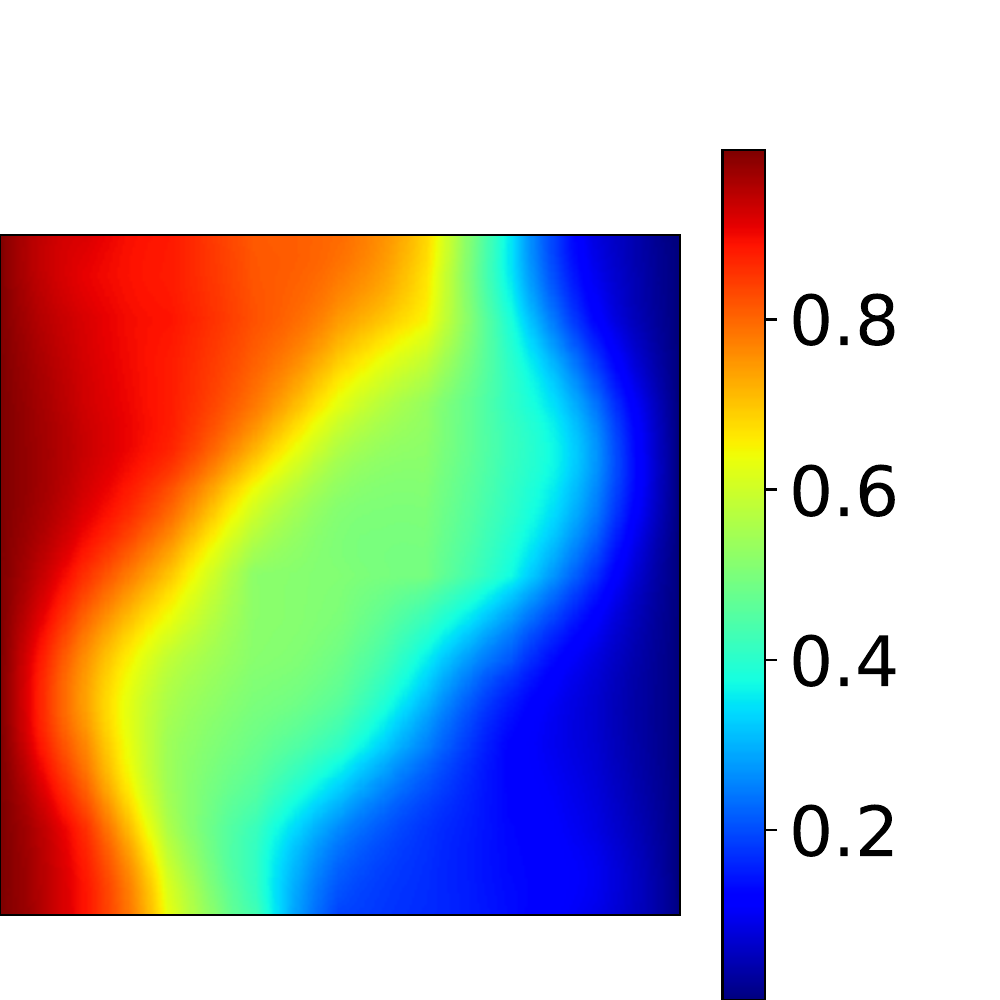}}&
		\tabincell{c}{\includegraphics[width=0.2\linewidth,trim={0 0 0 0},clip]{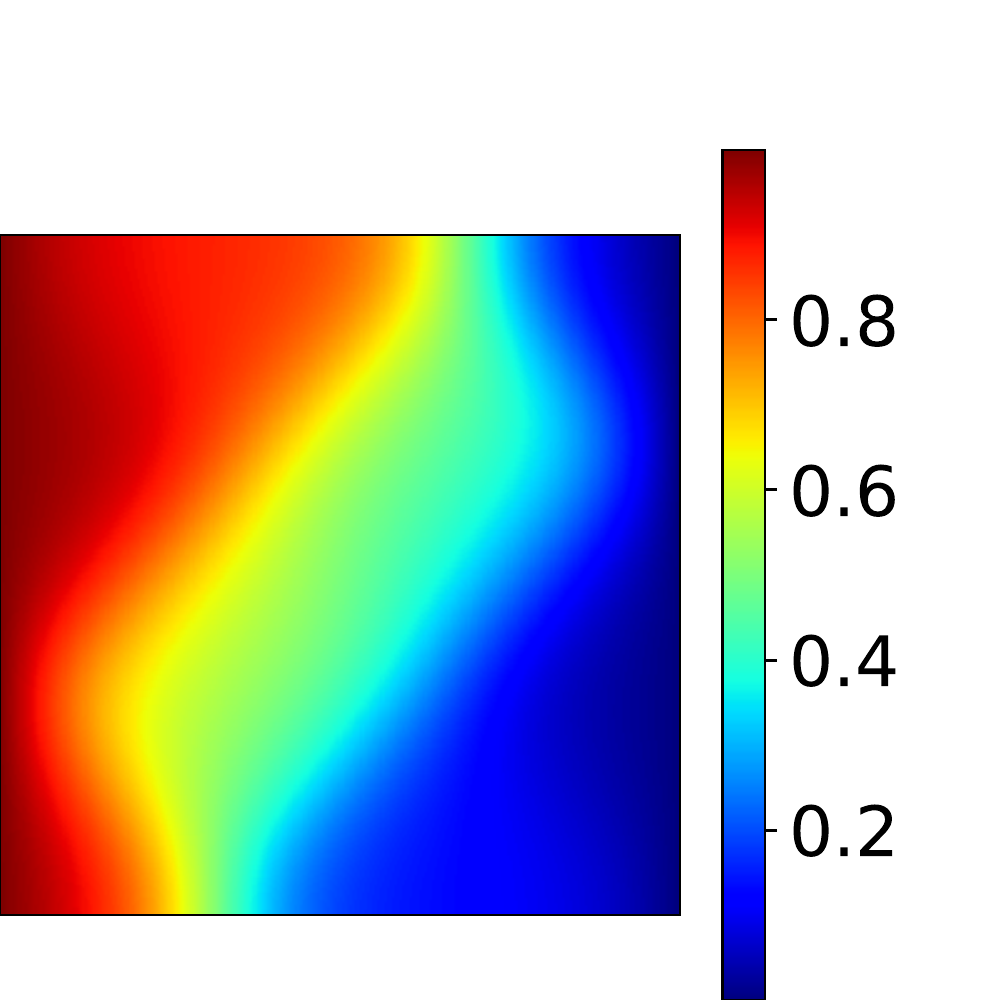}}&		\tabincell{c}{\includegraphics[width=0.2\linewidth,trim={0 0 0 0},clip]{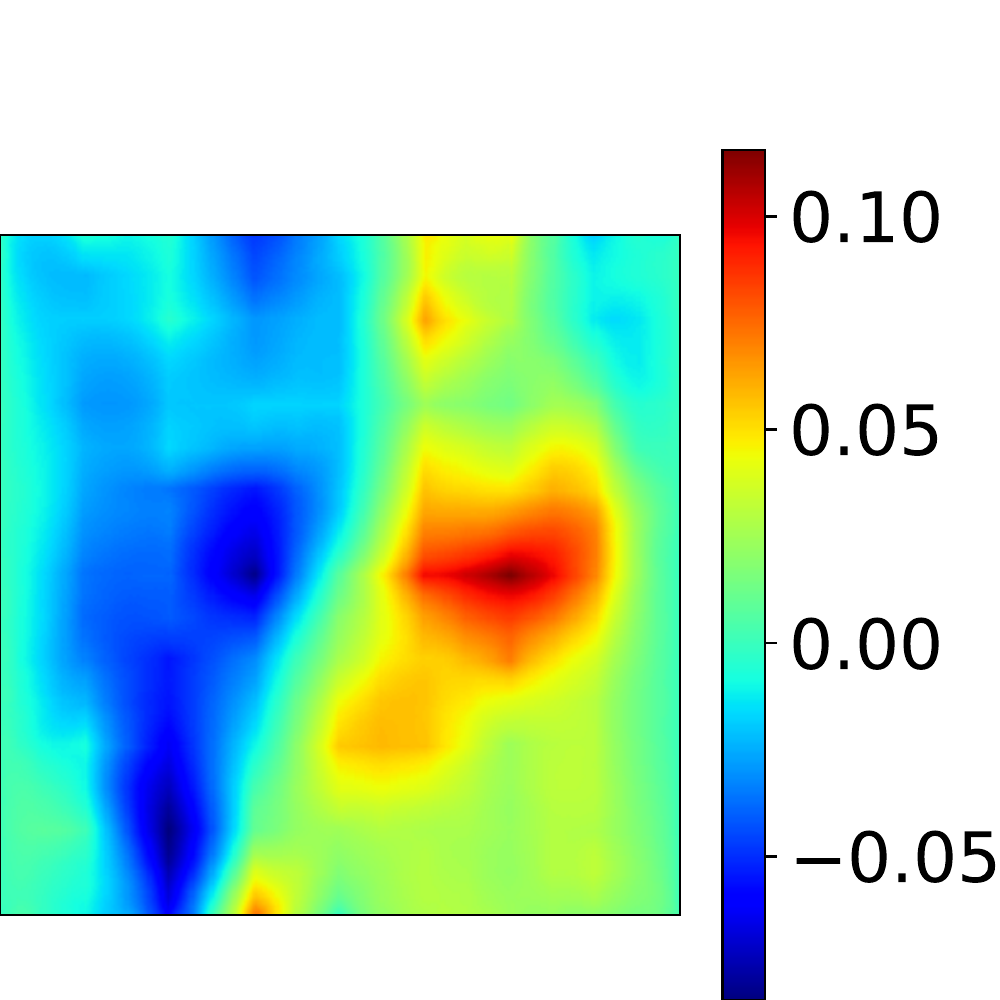}}\\
		&  \multicolumn{2}{c}{$ \underline{\omega} = (0.3105,  1.5386,  0.0932, -1.2442) $} &\\
		\hline 
		\tabincell{c}{\includegraphics[width=0.2\linewidth,trim={0 0 0 0},clip]{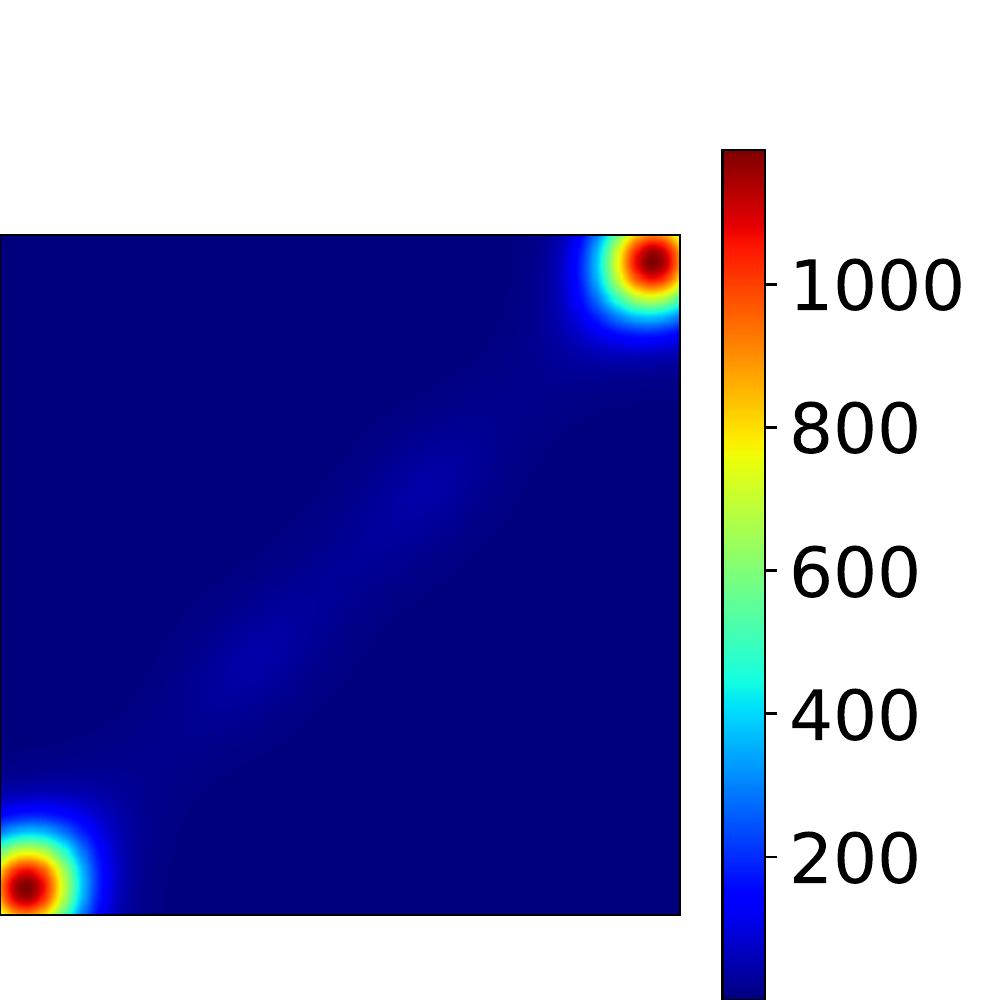}}&
		\tabincell{c}{\includegraphics[width=0.2\linewidth,trim={0 0 0 0},clip]{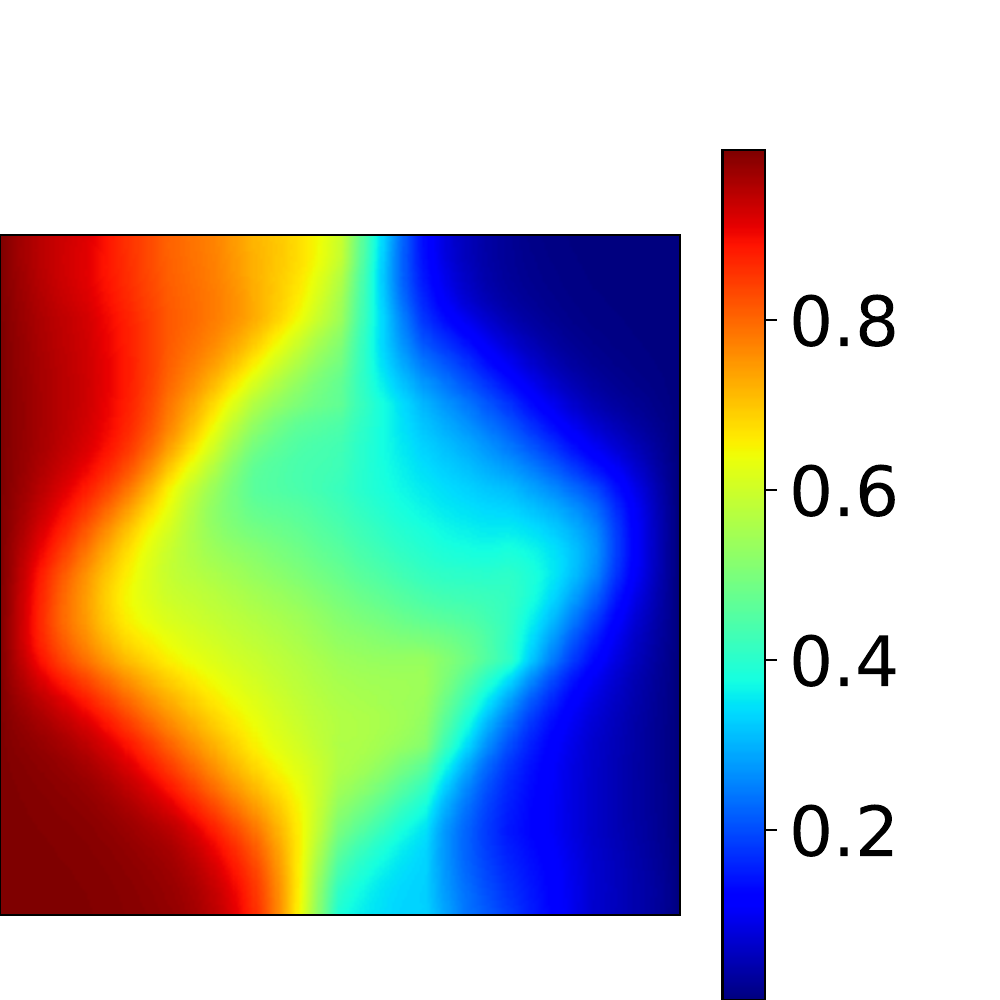}}&
		\tabincell{c}{\includegraphics[width=0.2\linewidth,trim={0 0 0 0},clip]{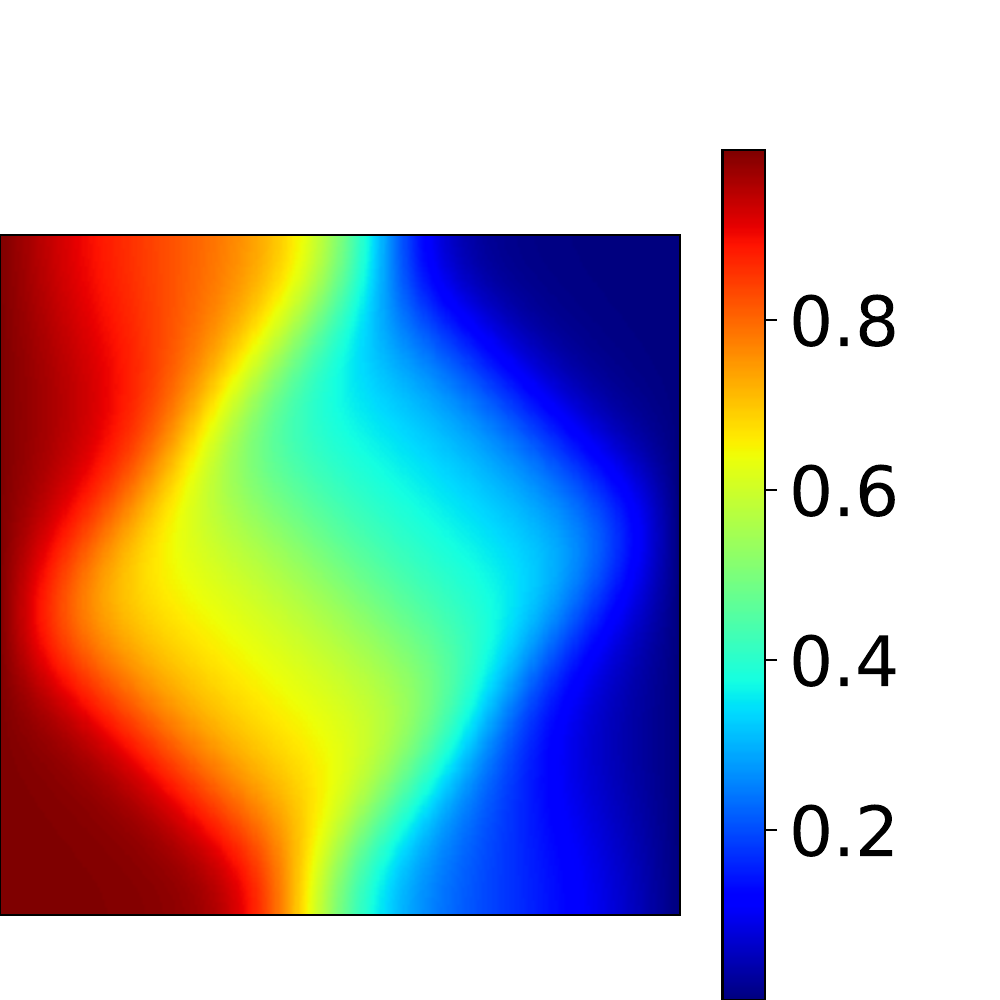}}&		\tabincell{c}{\includegraphics[width=0.2\linewidth,trim={0 0 0 0},clip]{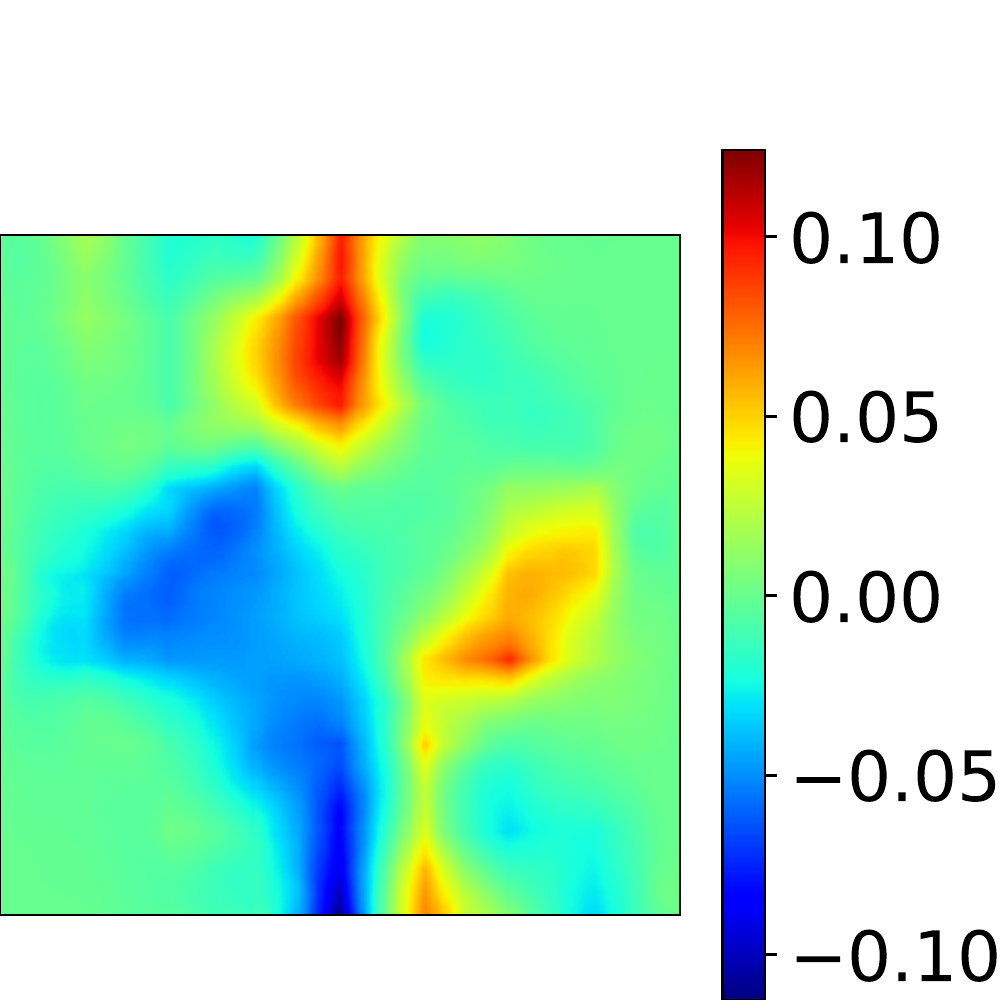}}\\
		&  \multicolumn{2}{c}{$ \underline{\omega} = (0.2838, -2.3550,  2.9574, -1.8963) $} &\\
		\hline 
		\tabincell{c}{\includegraphics[width=0.2\linewidth,trim={0 0 0 0},clip]{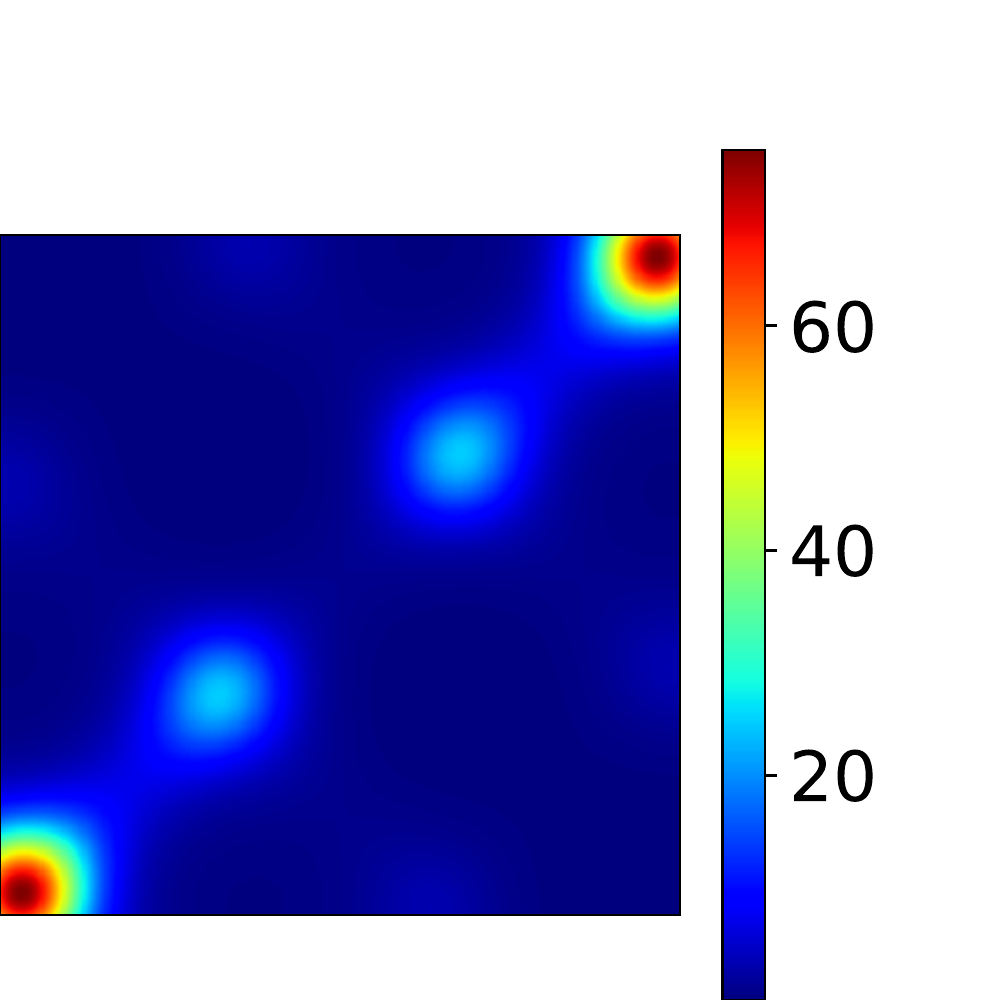}}&
		\tabincell{c}{\includegraphics[width=0.2\linewidth,trim={0 0 0 0},clip]{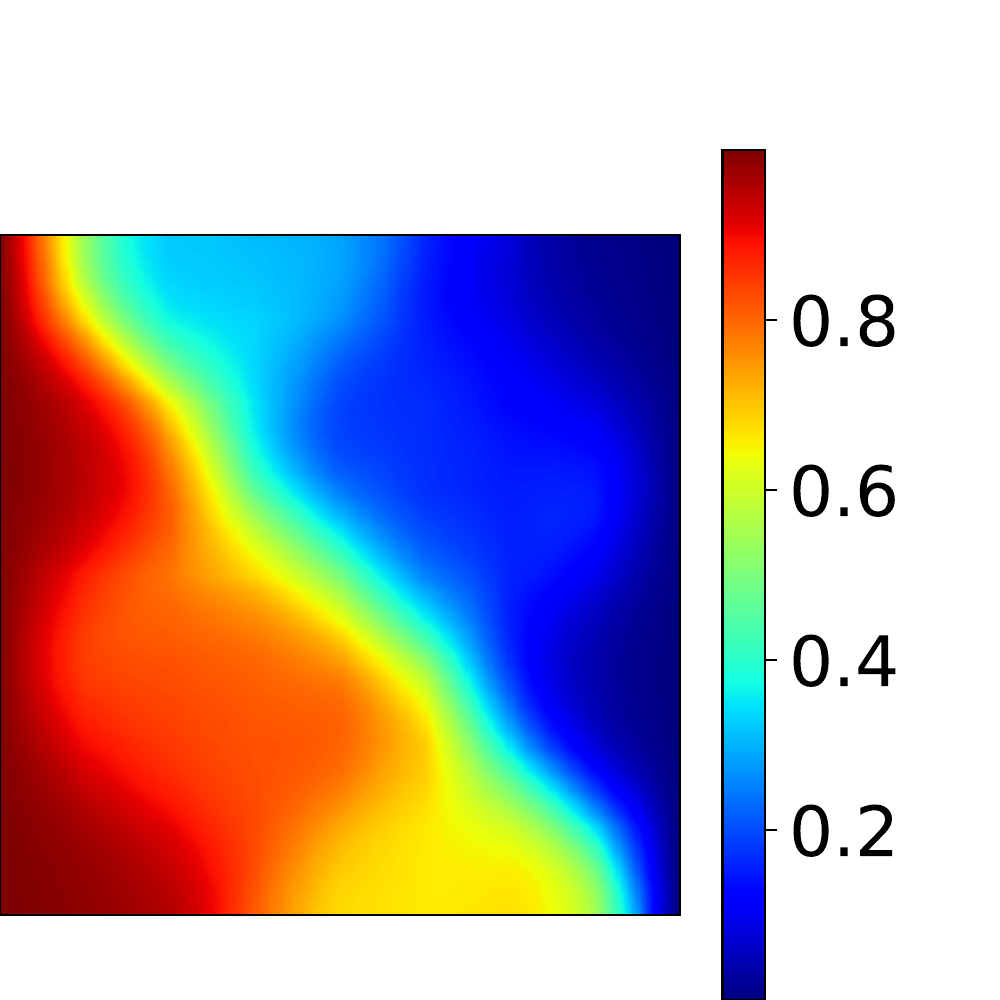}}&
		\tabincell{c}{\includegraphics[width=0.2\linewidth,trim={0 0 0 0},clip]{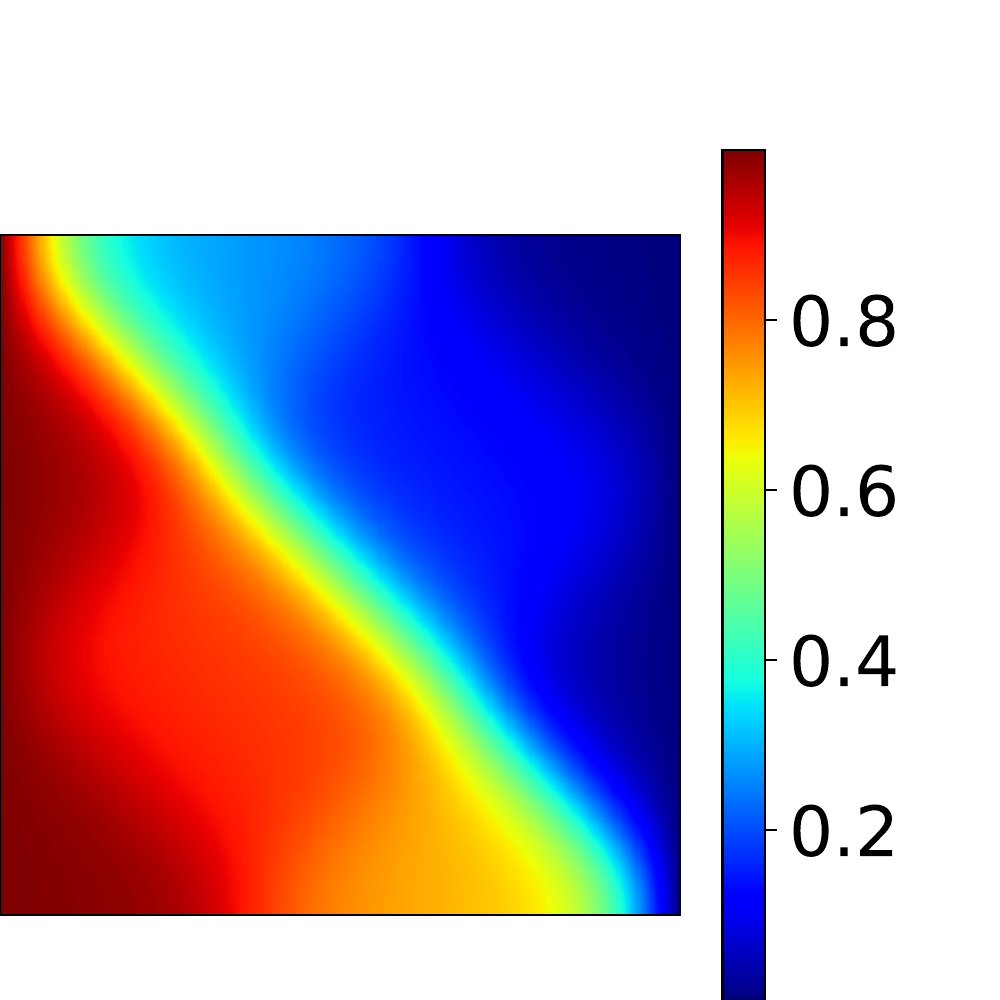}}&		\tabincell{c}{\includegraphics[width=0.2\linewidth,trim={0 0 0 0},clip]{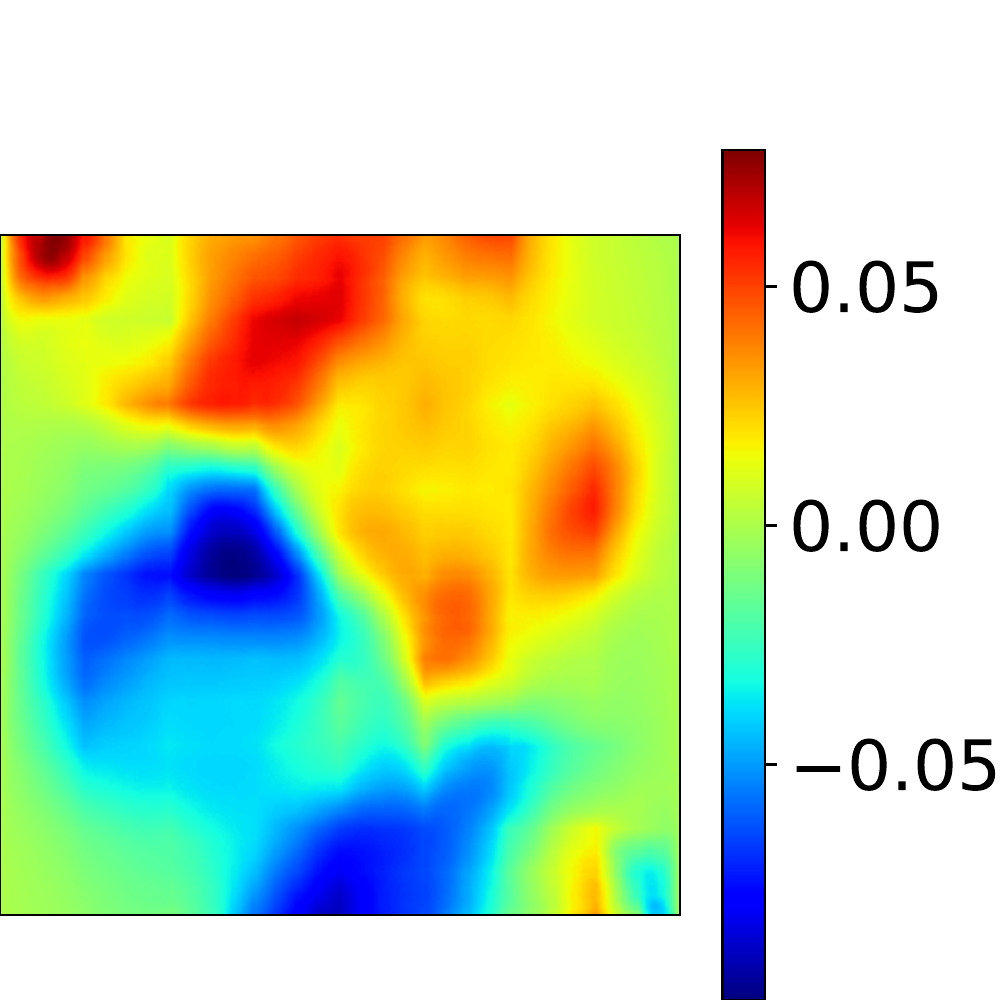}}\\
		&  \multicolumn{2}{c}{$ \underline{\omega} = (0.0293, -2.0943,  0.1386, -2.3271) $} &\\
		\hline 
    \end{tabular}
\end{table*}

\end{document}